\pdfoutput=1
\documentclass{article}
\usepackage[numbers]{natbib}

\usepackage[final]{neurips_2024_dataset}

\usepackage[utf8]{inputenc} 
\usepackage[T1]{fontenc}    
\usepackage{hyperref}       
\usepackage{url}            
\usepackage{booktabs}       
\usepackage{amsfonts}       
\usepackage{nicefrac}       
\usepackage{microtype}      
\usepackage{xcolor}         
\usepackage{multirow}
\usepackage{graphicx}
\usepackage{svg}
\usepackage{float}

\definecolor{ForestGreen}{rgb}{0, 0.69, 0.31}
\definecolor{NavyBlue}{rgb}{0, 0.44, 0.75}

\usepackage{xspace}
\providecommand{\eg}{\textit{e.g.}\@\xspace}

\newcommand{\etc}{\emph{etc.}\xspace}
\newcommand{\benchmarkname}{MMDU\xspace}
\newcommand{\instructname}{MMDU-45k\xspace}

\newcommand{\hgreen}[1]{\textcolor{ForestGreen}{\textbf{#1}}} 
\newcommand{\hblue}[1]{\textcolor{NavyBlue}{\textbf{#1}}} 

\definecolor{firstBest}{rgb}{0.9, 1, 0.9}
\definecolor{secondBest}{rgb}{1, 0.95, 0.95}

\title{\benchmarkname: A \underline{M}ulti-Turn \underline{M}ulti-Image \underline{D}ialog \underline{U}nderstanding Benchmark and Instruction-Tuning Dataset for LVLMs}

\author{%
  Ziyu Liu$^{1,2}$, Tao Chu$^{2}$, Yuhang Zang$^{\dagger2}$, Xilin Wei$^{2}$, Xiaoyi Dong$^{2}$, Pan Zhang$^{2}$, \\
  \textbf{Zijian Liang$^{1}$}, \textbf{Yuanjun Xiong$^{5}$}, \textbf{Yu Qiao$^{2}$}, \textbf{Dahua Lin$^{2,3,4}$}, \textbf{Jiaqi Wang$^{\dagger2}$} \\
  $^{1}$ SJTU, $^{2}$ Shanghai AI Laboratory, $^{3}$CUHK, $^{4}$ CPII under InnoHK, $^{5}$ MThreads, Inc.\\
  {\tt\small liuziyu77@sjtu.edu.cn, \{zangyuhang, wangjiaqi\}@pjlab.org.cn} \\
  {{\tt\small Github: \url{https://github.com/Liuziyu77/MMDU}}}
}

\begin{document}

\maketitle

\begin{abstract}
Generating natural and meaningful responses to communicate with multi-modal human inputs is a fundamental capability of Large Vision-Language Models (LVLMs). While current open-source LVLMs demonstrate promising performance in simplified scenarios such as single-turn single-image input, they fall short in real-world conversation scenarios such as following instructions in a long context history with multi-turn and multi-images. Existing LVLM benchmarks primarily focus on single-choice questions or short-form responses, which do not adequately assess the capabilities of LVLMs in real-world human-AI interaction applications.
Therefore, we introduce \textbf{\benchmarkname}, a comprehensive benchmark, and \textbf{\instructname}, a large-scale instruction tuning dataset, designed to evaluate and improve LVLMs' abilities in multi-turn and multi-image conversations.
We employ the clustering algorithm to find the relevant images and textual descriptions from the open-source Wikipedia and construct the question-answer pairs by human annotators with the assistance of the GPT-4o model.
\benchmarkname has a maximum of 18k image+text tokens, 20 images, and 27 turns, which is at least 5$\times$ longer than previous benchmarks and poses challenges to current LVLMs.
Our in-depth analysis of 15 representative LVLMs using \benchmarkname reveals that open-source LVLMs lag behind closed-source counterparts due to limited conversational instruction tuning data.
We demonstrate that fine-tuning open-source LVLMs on \instructname significantly addresses this gap, generating longer and more accurate conversations, and improving scores on \benchmarkname and existing benchmarks (MMStar: +1.1\%, MathVista: +1.5\%, ChartQA: +1.2\%). Our contributions pave the way for bridging the gap between current LVLM models and real-world application demands.  This project is available at \url{https://github.com/Liuziyu77/MMDU}.
\end{abstract}
\section{Introduction}

\begin{figure}[t]
  \centering
  \includegraphics[width=1.\linewidth]{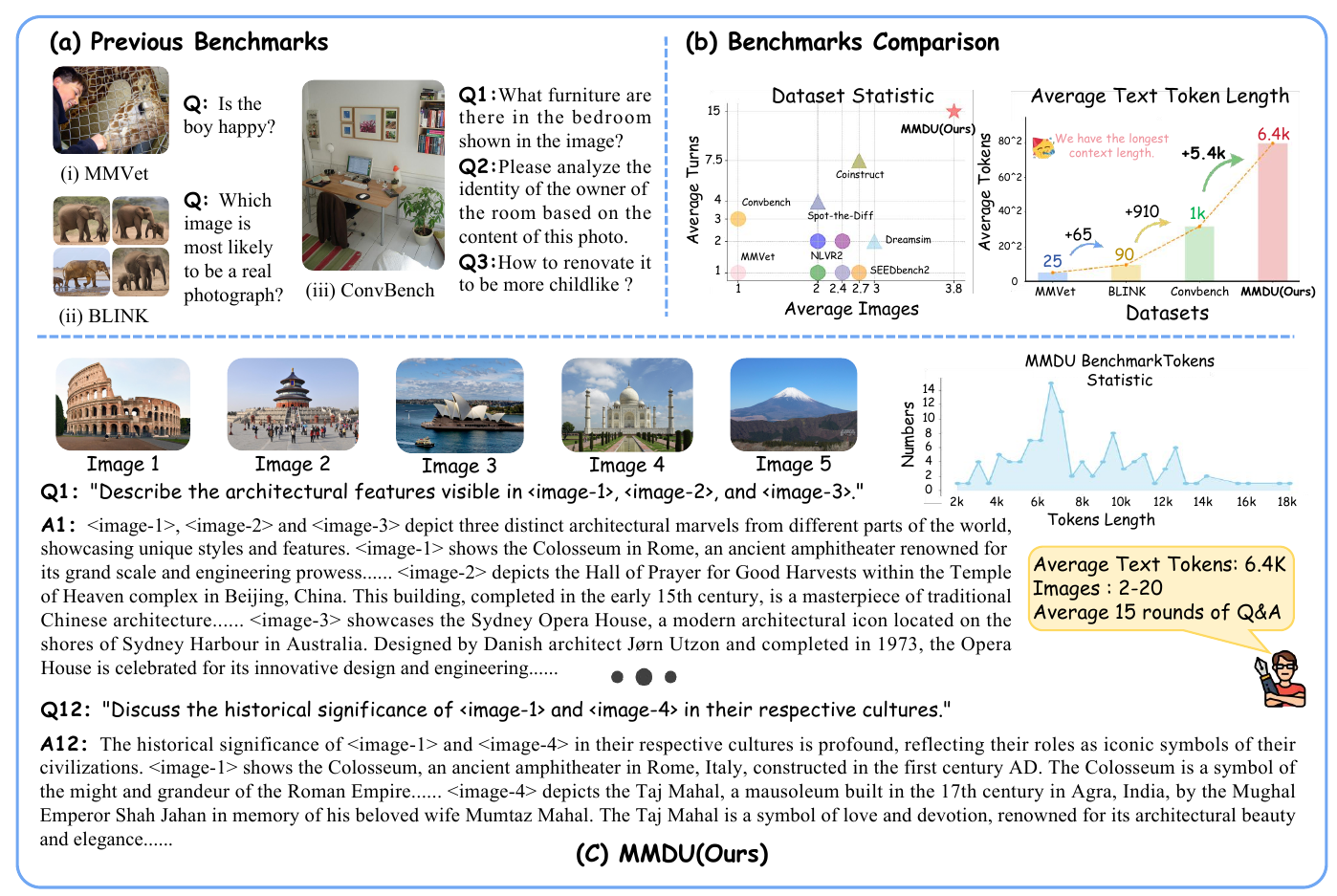}
  \vspace{-12pt}
  \caption{\textbf{Comparing \benchmarkname with previous LVLM benchmarks.} Our \benchmarkname (bottom) incorporates multi-turn and multi-image inputs, long context, and open-ended question-answering evaluation, making it more challenging and comprehensive than previous benchmarks (upper left).
  }
  \label{fig:dataset_compare}

\end{figure}

Human-AI interaction is a fundamental task to ensure that AI can be deployed in the real world for everyone, enabling inclusive and effective communication between humans and AI in various aspects of daily life. Current Large Vision-Language Models (LVLMs)~\cite{openai2023gpt4} have made significant strides in understanding and generating text conditioned on visual inputs, showing promising directions in AI assistant applications.

Current open-source LVLMs primarily focus on single-turn, single-image inputs, which are far from the complexities of real-world scenarios. In contrast, effective human-AI interaction in daily life demands a range of essential skills, including the ability to engage in multi-turn conversations that involve multiple image inputs and comprehend long-context histories to facilitate coherent and contextually appropriate conversations.
While existing benchmarks partially assess these abilities~\cite{liu2024convbench,suhr2018corpus,yu2023mm} (Fig.~\ref{fig:dataset_compare} \textbf{(a)}), they have limitations such as limited number of total tokens and do not provide a complete picture of a model's human-AI interaction capabilities. More challenging and comprehensive benchmarks are necessary to evaluate and advance these skills.

We present \textbf{\benchmarkname}, a comprehensive benchmark for multi-turn multi-image dialog understanding. Our data collection pipeline automatically selects relevant images and text descriptions from open-source Wikipedia~\cite{srinivasan2021wit}, forming the basis for multi-turn dialogues. We employ a clustering algorithm to identify relevant Wikipedia entities and design prompt templates for GPT-4o to generate multi-turn questions. Human annotators assess and refine GPT-4o's responses, producing ground-truth answers for our benchmark.

Our \benchmarkname benchmark possesses the following distinctive features: (1) \textbf{Multi-turn and Multi-image}: Our benchmark showcases a conversational setting with a maximum of 20 images and 17 turns, thereby surpassing the scope of preceding works (see Fig.~\ref{fig:dataset_compare} \textbf{(b)} and authentically replicating real-world chat assistant interactions. (2) \textbf{Long Context}: With a maximum of 18k text+image tokens, our benchmark evaluates the capacity of LVLMs to process and comprehend extended contextual information with a long context history. (3) \textbf{Open-ended Evaluation}: Departing from traditional benchmarks that rely on close-ended questions with concise outputs (\eg, multiple-choice questions or short answers), our benchmark adopts a more realistic and nuanced approach, assessing LVLM's performance through free-form multi-turn outputs that prioritize scalability and explainability, inspired by NLP research that leverages strong LLMs as judges~\cite{zheng2024judging}.

We evaluate 15 proprietary and open-source LVLMs on our \benchmarkname benchmark. Our evaluation reveals a significant performance disparity between proprietary and open-source LVLMs. The best open-source model scores 42.8\%, far behind the proprietary GPT-4o at 70.2\%. Notably, our observations provide a clear direction for improving the open-source models on long-context, multi-turn, and multi-image scenarios to bridge the performance gap. Based on our findings from the benchmark results on \benchmarkname, the practical need urges the visual instruction tuning data containing multi-turn and multi-images for open-source LVLMs.

To get one step closer to proprietary LVLM models, we further present \textbf{\instructname}. We collect \textbf{45k} high-quality instruction tuning data using the same process employed in building \benchmarkname, with a random sampling of human verification instead of the exhaustive human evaluation used in \benchmarkname.
Adding our instruction tuning data \instructname into the LVLM supervised fine-tuning (SFT) stage improves performance on various benchmarks, such as boosting InternLM-XC2~\cite{dong2024internlm}'s performance by $14.5\%$/$1.1\%$/$1.5\%$/$1.2\%$ on MMDU/MMStar~\cite{chen2024we}/MathVista~\cite{lu2024mathvista}/ChartQA~\cite{masry2022chartqa}, respectively.

Our main contribution is summarized: \textbf{(1)} We introduce \benchmarkname that assesses the multi-turn, multi-image dialog understanding capabilities of LVLMs, specifically designed for human-AI interaction.  \textbf{(2)}  We conduct a comprehensive evaluation of existing LVLMs on \benchmarkname, revealing significant challenges in this task and providing valuable insights for future LVLM development. \textbf{(3)} We present \instructname, a large-scale instruction tuning dataset designed to enhance dialog understanding abilities. We demonstrate that fine-tuning LVLMs on \instructname leads to improved performance on both \benchmarkname and existing benchmarks.

\section{\benchmarkname Benchmark}

\subsection{Benchmark Overview}
Although many LVLMs now claim to handle tens of thousands, hundreds of thousands, or even millions of tokens in length, their actual performance significantly declines in real-world applications as the number of images or the length of the context increases. Both the dialogue quality and image recognition capabilities of LVLMs deteriorate notably under these conditions.

To evaluate the multi-image multi-turn dialogue capabilities of existing models, we have developed the \benchmarkname Benchmark. Our benchmark comprises 110 high-quality multi-image multi-turn dialogues with more than 1600 questions, each accompanied by detailed long-form answers. Previous benchmarks typically involved only single images or a small number of images, with fewer rounds of questions and short-form answers. However, \benchmarkname significantly increases the number of images, the number of question-and-answer rounds, and the in-context length of the Q\&A. The questions in \benchmarkname involve 2 to 20 images, with an average image\&text token length of 8.2k tokens, and a maximum image\&text length reaching 18K tokens, presenting significant challenges to existing multimodal large models. For more data statistics about MMDU, please refer to Tab.\ref{tab:statistics} and Fig.\ref{fig:document_distribution}.

\benchmarkname aims to test models' abilities to simultaneously understand multiple images and follow instructions in long dialogues. We design precise prompts to evaluate the models' responses, and our evaluation criteria details are discussed in Sec.~\ref{sec:evaluation}.

\subsection{Benchmark Construction}
\textbf{Data Collection.}\label{sec:data_collection}
Our goal in constructing this benchmark is to measure the current models' ability to understand multiple images and generate long texts in general scenarios.

The first step is selecting appropriate multiple images and related textual information as the foundation for multi-turn dialogues. Given that the generated dialogue content needs to be logically coherent and rich in content, we cannot use random sets of images to build the Q\&A pairs. Random images would lead to low-quality and illogical dialogues, both in the question-construction and answer-generation processes.

\begin{figure}[t]
  \centering
  \includegraphics[width=1.\linewidth]{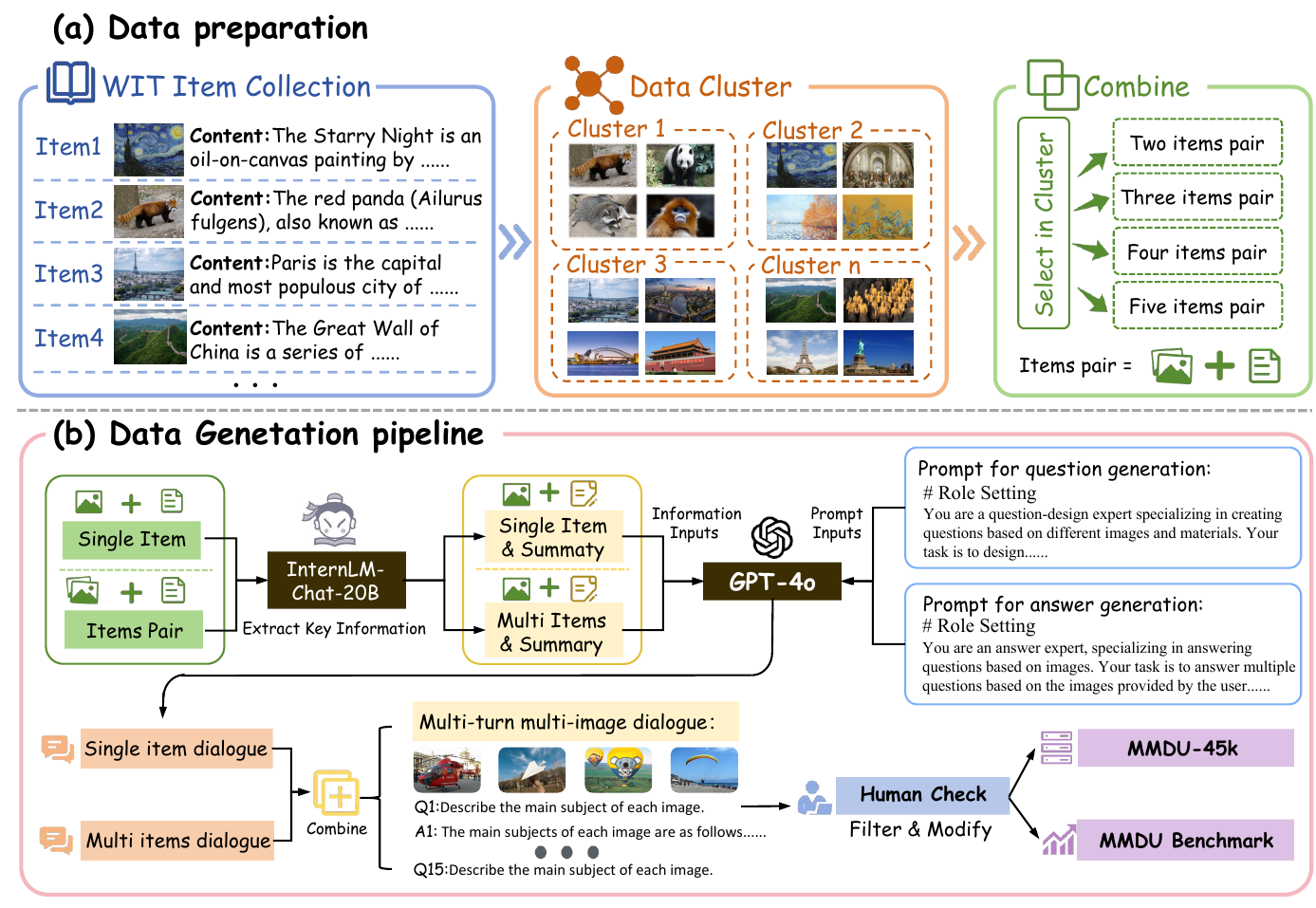}
  \vspace{-12pt}
  \caption{\textbf{An overview of \textbf{(a)} data preparation and \textbf{(b)} generation pipeline for \benchmarkname and \instructname}. We first collect the relevant image and text descriptions from Wikipedia using the clustering algorithm. Then we prompt GPT-4o to design multi-turn questions. The human annotators revise the GPT-4o response as the ground-truth answers.}
  \label{fig:pipeline}
\end{figure}

To address this issue, we employ a clustering method to construct high-quality image sets. We extensively screened entries on the open-source Wikipedia~\cite{srinivasan2021wit}, encoded the relevant tags of entries using a sentence transformer\cite{reimers-2019-sentence-bert}, and clustered the entries using the obtained embeddings. After clustering enough entries of the same category together, we further matched them using image captions to obtain highly relevant entries and image sets. Then, within each cluster, we selected multiple images and their associated textual information to create combinations of image-text pairs, ranging from 2 to 20 images. The process of collecting and clustering entries is illustrated in Fig.~\ref{fig:pipeline} \textbf{(a)}.

\textbf{Construction with GPT-4o.} After obtaining the combinations of multiple images, we use carefully crafted prompts to guide the GPT-4o model in generating corresponding questions and answers based on the available images and text information. Initially, we constructed multi-turn Q\&A pairs for each single image and its associated text. Then, we input the combinations of multiple images into GPT-4o to generate multi-turn Q\&A pairs based on multiple images, ensuring through prompts that the questions covered multiple different images simultaneously.

Building on this, we combined the multi-turn Q\&A pairs for multiple images with those for each individual image, creating dialogues that include both single-image and multi-image questions.
To ensure the quality of the benchmark, we invited experts to meticulously review the generated dialogues, selecting 110 high-quality multi-turn, multi-image dialogues for our benchmark. Additionally, we carefully edited these 110 samples to eliminate hallucinations and errors in GPT-4o's responses, ensuring the accuracy and richness of the benchmark content. Our pipeline is shown in Fig.~\ref{fig:pipeline} \textbf{(b)}.

Furthermore, our generated multi-turn, multi-image data is highly scalable. During the Q\&A construction process, we required GPT-4o to organize the generated text according to our specified Text-Image Interleaving Format, using tags like <image-1>, <image-2>, \etc, to refer to different images. Our design is flexible to treat the generated multi-turn, multi-image dialogues as fundamental components. By modifying the values in <image-$i$>, we can concatenate multiple dialogues, thereby constructing dialogues involving dozens or even hundreds of images. Our data is not limited to a few images per Q\&A generation but is capable of supporting dialogues of theoretically unlimited length.

\begin{figure}[t]
  \centering
  \includegraphics[width=1.\linewidth]{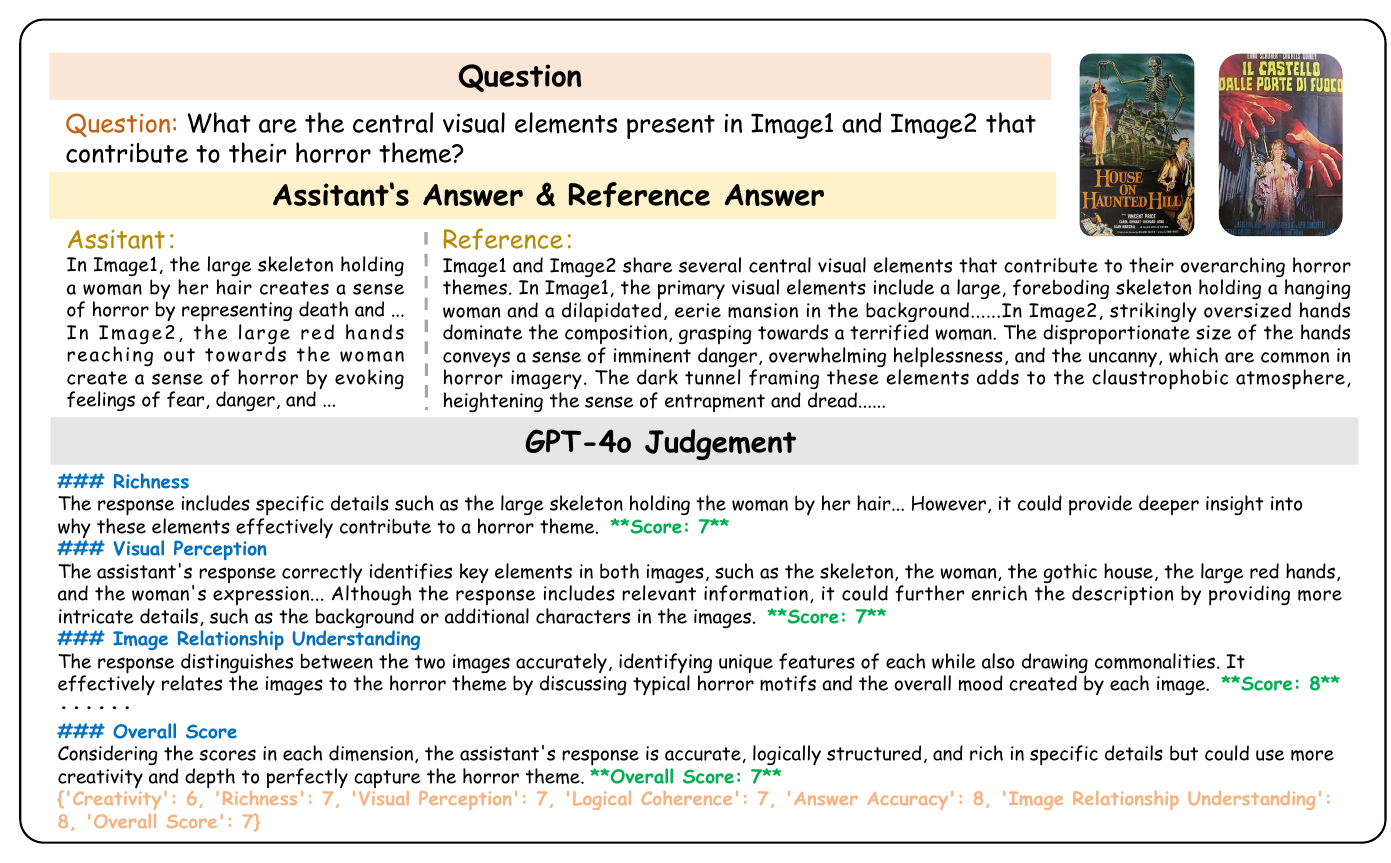}
  \vspace{-12pt}
  \caption{
  \footnotesize
  \textbf{The evaluation pipeline of \benchmarkname}. We use the GPT-4o as a judge to give the overall score based on the referenced answer. In each evaluation, GPT-4o will refer to both the model's answer and the reference answer. It will provide corresponding scores (in green) for each evaluation criterion (in blue), and finally, summarize the results (in light orange).
  }
  \label{fig:llm_as_judge}
\end{figure}

\paragraph{Quality Control with Human Annotators}

In the process of constructing the dataset, we implemented two stringent measures to ensure its quality: \textbf{(1)} We combined automated and manual screening methods to select images and texts that meet our standards. Specifically, we performed an initial screening using clustering techniques on a large-scale image and text database, automatically removing low-quality, blurry, or irrelevant images and texts. This ensured that the image combinations and their corresponding texts were of high quality and relevance. \textbf{(2)} To avoid hallucinations and errors in the model-generated dialogues, we enforced strict quality control on the texts generated by GPT-4o. We introduced a multi-round manual review mechanism. Each set of Q\&A underwent at least two rounds of manual review: the first round involved preliminary checks by regular reviewers, and the second round involved in-depth examination and modification by experts. 

For the  two rounds of manual review, our experts reviewed and corrected (by removing or rewriting) any hallucinations and errors to ensure that all dialogues are accurate. To facilitate verification, we designed a specialized web UI that allows for quick browsing and modification of data content. Please refer to the Appendix ~\ref{appendix:webui} to check the interface of our web UI used for the human check process. 

During the data annotation process, all our annotators were either junior PhD-level students or senior researchers, with a total of 20 participants. Senior researchers or PhD students with relevant professional backgrounds were selected as experts. Before data annotation, all annotators underwent training, and a small subset of data was pre-annotated. Once the pre-annotation results met the required standards, the subsequent annotation process was carried out. 

Since our images are sourced from Wikipedia entries, each image is accompanied by a corresponding caption and all related information from the Wiki entry where the image is found. During the manual annotation process, annotators can easily understand the specific content and background information of each image by reading the Wiki entry. Therefore, there is no risk of introducing extra annotation errors due to misunderstanding of the image content. The various strategies mentioned above ensured that the final dataset was not only accurate but also of high academic and practical value.

\subsection{Evaluation}\label{sec:evaluation}



Evaluating the subjective, open-ended, free-form, and long-context visual question-answering pairs is indeed challenging. Traditional metrics (e.g., BLEU-4, CIDEr) often suffer from shortcomings like neglecting semantic understanding and failing to capture long-distance dependencies, they are not popular choices in recent days.

Inspired by NLP research that leverages strong LLMs as judges~\cite{zheng2024judging}, we have developed an evaluation pipeline using GPT-4o to evaluate model performance. Specifically, following the generation of model predictions on our benchmark dataset, GPT-4o evaluates these predictions across various dimensions for each turn and sample, comparing them against reference answers. The aggregated results across multiple turns are averaged to derive sample scores, and the average scores across all samples constitute our benchmark scores. This method excels at understanding context and semantics, providing more accurate evaluations of visual content, and capturing long-distance dependencies, which traditional metrics often miss.

To ensure a comprehensive and nuanced evaluation, we have identified six dimensions: Creativity, Richness, Visual Perception, Logical Coherence, Answer Accuracy, and Image Relationship Understanding. To guide GPT-4o in providing balanced and equitable assessments, we have meticulously crafted evaluation prompts for each dimension. Each dimension's score range of 10 is divided into five intervals (0-2, 2-4...8-10), with corresponding judgment criteria established for each interval. GPT-4o follows these criteria to conduct judgment processes and deliver final scores for each dimension. As illustrated in Fig~\ref{fig:llm_as_judge}, guided by our prompts, GPT-4o assesses the assistant's responses against reference answers, offering both a reasonable score and a transparent judgment process. Please refer to the Appendix~\ref{appendix:benchmark} for our judgment prompts.

\section{\instructname for Instruct Tuning}
 
 \subsection{Dataset Construction}
We follow the same process as constructing the benchmark to build our \instructname. First, we collect a vast number of Wikipedia entries and extracted tags from these entries, including wiki tree labels and image captions. We use sentence transformers to encode the textual information and then apply the clustering algorithm to obtain the embeddings. During the clustering process, we calculate the cosine similarity between different embeddings and group highly related entries into clusters by setting a threshold $\tau=0.75$.
From the clusters with high relevance, we select multiple images and their corresponding entry information and perform information extraction and filtering using InternLM-chat-20B\cite{cai2024internlm2}. We design precise prompts to guide GPT-4o in generating multi-image, multi-round, long dialogues based on the information filtered by InternLM-Chat-20B. 

During the dataset construction process, we obtain several clusters with a wide range of category distributions. This ensures that our dataset comprehensively covers various aspects of real life, including geography, history, culture, mathematics, physics, chemistry, animals, plants, food, and more. This rich knowledge will help LVLM learn long-context conversational abilities in general scenarios of the real world.

In the manual data inspection phase, due to the large volume of data in the \instructname, it was not feasible to review all of the data, so we sampled $5\%$ of the dataset for inspection. Statistical analysis showed that the probability of hallucinations and errors in this subset was less than $5\%$, indicating a high level of reliability.

\begin{minipage}{0.5\textwidth} 
 \begin{table}[H]
 \fontsize{8.2pt}{\baselineskip}\selectfont 
 \renewcommand\tabcolsep{2pt} 
 \renewcommand\arraystretch{0.8} 
 \centering
 \begin{tabular}{lc}
 \toprule
 \textbf{Statistic} & \textbf{Number} \\
 \midrule
 \textbf{\benchmarkname Benchmark} & 110 \\
  ~- Avg./Max. Image\&Text Tokens & 8.2k/18k \\
  ~- Avg./Max. Images & 3.8/20 \\
  ~- Avg./Max. Turns & 15/27 \\
  ~- Number of QA Pairs & 1645 \\
  \midrule
  \textbf{\instructname} & 45k \\
  ~- Avg./Max.  Image\&Text Tokens & 5k/17k \\
  ~- Avg./Max Images & 3/5 \\
  ~- Avg./Max. Turns & 9/27 \\
  ~- Number of QA Pairs & 410k \\
  ~- Single-image Related Questions & 40k \\
  ~- Multi-images Related Questions & 369k \\
  ~- Avg./Max. Question Length & 31/91 \\
  ~- Avg./Max. Answer Length & 368/1518 \\
 \bottomrule
 \end{tabular}
 \caption{\footnotesize \textbf{Statistics on \benchmarkname and \instructname.}}
 \label{tab:statistics}
 \end{table}
 \end{minipage}
 \hfill
 \begin{minipage}{0.5\textwidth}
 \begin{figure}[H]
 \centering
\includegraphics[width=.9\linewidth]{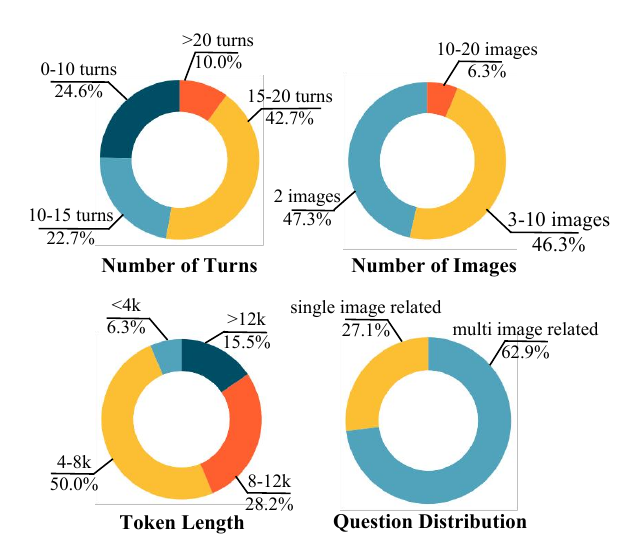}
\vspace{-6pt}
 \caption{\footnotesize \textbf{Detailed distribution of \benchmarkname.}}
 \label{fig:document_distribution}
 \end{figure}
 \end{minipage}

\subsection{Dataset Statistics}
In the \instructname, we construct a total of 45k instruct tuning data conversations. The data statistics are shown in Tab.~\ref{tab:statistics}.
Each data in our \instructname dataset features an ultra-long context, with an average image\&text token length of 5k and a maximum image\&text token length of 17k tokens. Each dialogue contains an average of 9 turns of Q\&A, with a maximum of 27 turns. Additionally, each data includes content from 2-5 images. The dataset is constructed in a well-designed format, providing excellent scalability. It can be expanded to generate a larger number and longer multi-image, multi-turn dialogues through combinations.
The image-text length and the number of turns in \instructname significantly surpass those of all existing instruct tuning datasets. This enhancement greatly improves the model's capabilities in multi-image recognition and understanding, as well as its ability to handle long-context dialogues.

\section{Experiments}\label{sec:experiments}
We evaluate previous representative LVLMs on our \benchmarkname benchmark in Sec.~\ref{sec:benchmark_resuls} and present the analysis of our findings.
To demonstrate the high quality of our instruction tuning data \instructname, we provide the comparison results of adding \instructname in the LVLM SFT stage in Sec.~\ref{sec:sft}.

\paragraph{Baselines}
We report the performance of four closed-source API models: QWen-VL-Max~\cite{bai2023qwen}, Claude3~\cite{claude3}, GPT-4-turbo~\cite{openai2023gpt4} and GPT-4o~\cite{gpt4o}. We also present the performance of 11 LVLMs including Monkey~\cite{li2023monkey}, Idefics2~\cite{laurençon2024matters}, LLaVa1.5 7B/13B~\cite{liu2023improved}, Deepseek-VL~\cite{lu2024deepseek}, MiniCPM-v-2.5~\cite{yu2024rlaifv,xu2024llava-uhd}, Yi-VL~\cite{young2024yi}, InternVL-Chat-V1.5~\cite{chen2024far}, InternLM-XC2~\cite{dong2024internlm}, Qwen-VL-7B~\cite{bai2023qwen} and LLaVa1.6~\cite{liu2024llavanext}. Please refer to the Appendix~\ref{appendix:evaluation} for the details of our baselines.

\subsection{Main Results on \benchmarkname}\label{sec:benchmark_resuls}
\begin{table*}[t]
    \footnotesize
    \begin{center}
    \caption{
    \footnotesize
    \textbf{Evaluation results of different LVLMs on \benchmarkname.} We report the metrics of Creativity (C), Richness (R), Visual Perception (VP), Logical Coherence (LC), Answer Accuracy (AA), Image Relationship Understanding (IRU), and the averaged (Avg.) results.
    }
    \vspace{+5pt}
    \label{tab:exps_mmdu}
    \renewcommand{\arraystretch}{0.5}
    \scalebox{1.}{
    \begin{tabular}{{@{}l c| c c c c c c |c  @{}}}
    \toprule
    Models & Param & C & R & VP & LC & AA & IRU & Avg.\\
    \midrule
    \multicolumn{9}{c}{\textit{Closed-source LVLMs}} \\
    \midrule
    Qwen-VL-Max~\cite{qwen1.5} & -  & 40.3 & 40.2 & 46.2 & 62.5 & 51.6 & 45.9 & 46.9\\
    Claude3-Opus~\cite{claude3} & -  & 58.6 & 61.5 & 59.7 & 75.1 & 64.1 & 59.8 & 62.6\\
    GPT-4-turbo~\cite{openai2023gpt4} & -  & 62.0 & 64.2 & 63.4 & 78.0 & 69.0 & 64.4 & 66.3 \\ 
    GPT-4o~\cite{gpt4o} & -  & 63.7 & 69.6 & 66.7 & 80.6 & 73.3 & 68.1 & 70.2\\
    \midrule
    \multicolumn{9}{c}{\textit{Open-source LVLMs}} \\
    \midrule
    Monkey~\cite{li2023monkey} & 10B & 11.9 & 12.0 & 14.8 & 21.9 & 19.6 & 14.6 & 14.1\\
    Idefics2~\cite{laurençon2024matters} & 8B & 17.8 & 17.6 & 27.9 & 43.1 & 32.8 & 26.9 & 25.4\\
    LLaVa1.5-7B~\cite{liu2023improved} & 7B  & 27.8 & 28.0 & 33.2 & 43.0 & 35.4 & 31.7 & 32.2\\
    Deepseek-VL~\cite{lu2024deepseek} & 8B  & 27.3 & 27.7 & 31.2 & 38.7 & 33.2 & 30.0 & 30.8\\
    MiniCPM-v-2.5~\cite{yu2024rlaifv,xu2024llava-uhd} & 8B  & 27.0 & 26.4 & 33.2 & 48.9 & 38.6 & 32.2 & 33.0\\
    Yi-VL~\cite{young2024yi} & 6B  & 31.7 & 32.2 & 30.6 & 47.5 & 34.0 & 30.0 & 33.2\\
    LLaVa1.5-13B~\cite{liu2023improved} & 13B  & 31.5 & 31.2 & 35.1 & 46.2 & 38.1 & 34.3 & 35.3\\
    InternVL-Chat-V1.5~\cite{chen2024far} & 26B  & 31.2 & 31.5 & 37.4 & 52.6 & 41.7 & 36.1 & 37.4\\
    InternLM-XC2~\cite{dong2024internlm} & 7B & 29.7 & 29.5 & 36.2 & 50.1 & 40.3 & 35.2 & 35.6\\
    Qwen-VL-7B~\cite{bai2023qwen} & 7B & 33.4 & 33.6 & 39.2 & 53.8 & 43.1 & 38.1 & 39.3\\   
    LLaVa1.6-mistral~\cite{liu2024llavanext} & 7B & 37.7 & 39.3 & 41.4 & 57.2 & 45.6 & 40.2 & 42.8\\   
    \cmidrule(lr){1-1} \cmidrule(lr){2-2} \cmidrule(lr){3-8} \cmidrule(lr){9-9}
    LLaVa 1.5~\cite{liu2023improved} + \instructname & 7B & 34.3 & 34.5 & 36.7 & 47.2 & 38.5 & 35.5 & 37.2\\
    $\Delta$ &  & \hgreen{+6.5} &\hgreen{+6.5} &	\hgreen{+3.5} &	\hgreen{+4.2} &	\hgreen{+3.1} &	\hgreen{+3.8} &	\hgreen{+5.0}\\
    \cmidrule(lr){1-1} \cmidrule(lr){2-2} \cmidrule(lr){3-8} \cmidrule(lr){9-9}
    InternLM-XC2~\cite{dong2024internlm} + \instructname & 7B & 45.6 & 43.9 & 49.9 & 64.1 & 53.0 & 48.7 & 50.1\\
    $\Delta$ &  & \hgreen{+15.9} & \hgreen{+14.4} & \hgreen{+13.7} & \hgreen{+14.0} & \hgreen{+12.7} & \hgreen{+13.5} & \hgreen{+14.5} \\
    \bottomrule     
    \end{tabular}}
    \end{center}
\end{table*}
Tab.~\ref{tab:exps_mmdu} presents the benchmarking results on our \benchmarkname benchmark. Our key findings are summarized as follows. (1) Our benchmark poses significant challenges to current LVLMs. Notably, even the advanced GPT-4o model achieves an average accuracy of only 70.2\%, while open-source LVLMs achieve merely 42.8\% or lower, indicating substantial room for improvement. (2) We observe a significant performance gap between closed-source LVLMs and open-source LVLMs. We speculate that this disparity arises from the scarcity of open-source instruction tuning data with multi-turn and multi-image capabilities, leading to limited improvement in open-source LVLMs. This inspired us to collect and release \instructname, a valuable resource for the open-source community, to bridge this gap. 

For Tab.~\ref{tab:exps_mmdu}, we found that InternLM-Xcomposer2 benefits more from MMDU than LLaVA1.5. This is because, as a more recent model, InternLM-Xcomposer2 uses a different LLM backbone, a more powerful architecture, larger pre-training and SFT data, and more advanced training strategies compared to LLaVa1.5. These advantages may provide InternLM-Xcomposer2 with stronger generalization capabilities.

In addition, we conduct experiments to evaluate the quality of our evaluation with GPT4-o by comparing it to human judgment. Specifically, experts score the results predicted by each model on our benchmark using the same judgment criteria as our evaluation. We calculated several similarity metrics for the overall scores between experts and the GPT4-o system. The Pearson similarity of 97.5\% indicates a strong linear relationship, while the Spearman similarity of 97.3\% demonstrates consistent scoring monotonicity. The Kendall similarity of 89.0\% suggests some variability in the manual scores compared to the judgment range of GPT4-o, yet the consistency remains high.

\subsection{Fine-tuning Results using \instructname}\label{sec:sft}
\begin{figure}[t]
  \centering
  \includegraphics[width=1.\linewidth]{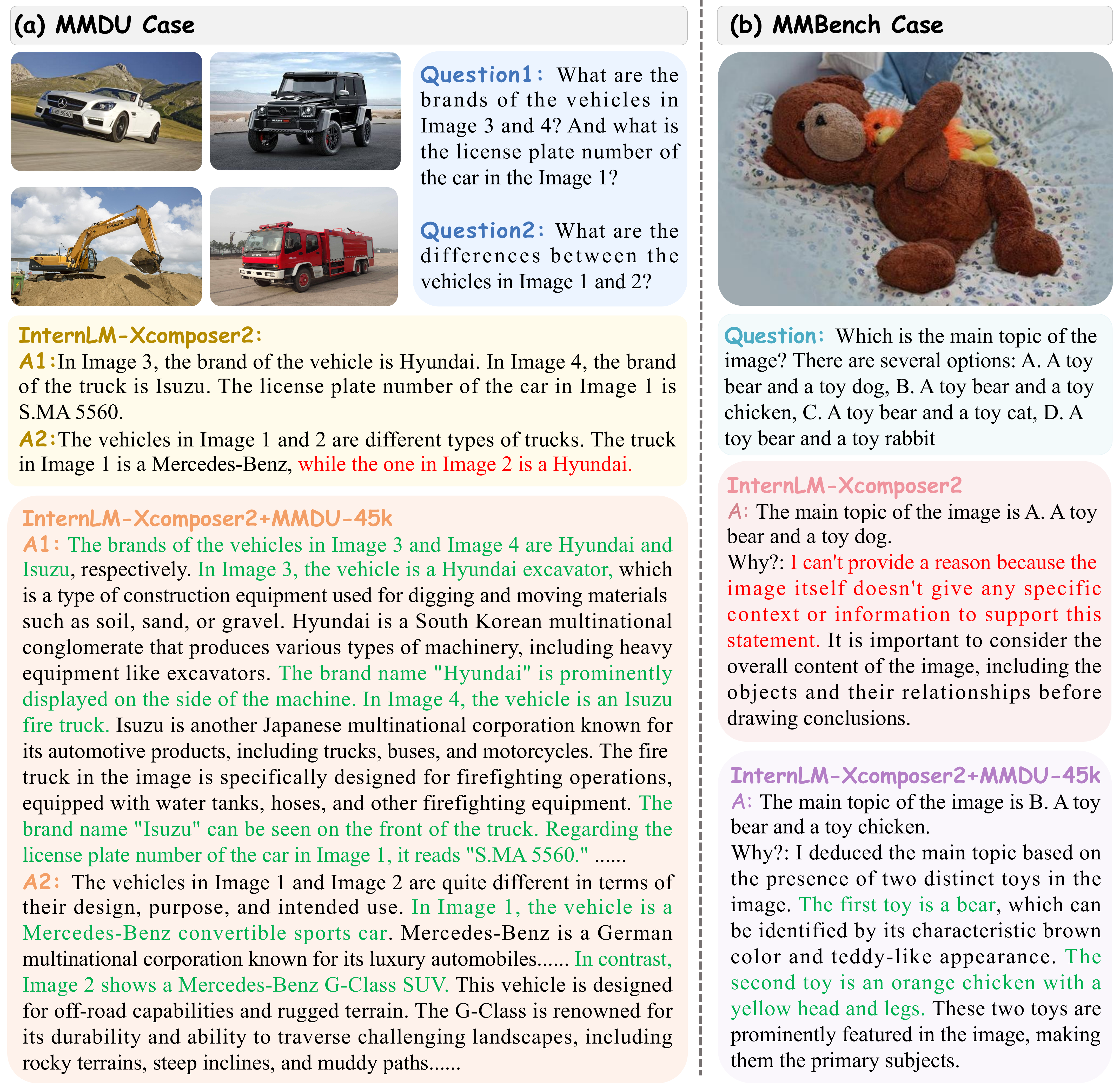}
  \vspace{-16pt}
  \caption{\textbf{Visualization examples} of adding \instructname in the LVLM SFT stage. Error/hallucination descriptions are marked in \textbf{\textcolor{red}{red}}, and the detailed and accurate descriptions are marked in \textbf{\textcolor{ForestGreen}{green}}. The case on the left is from MMDU, and the case on the right is from MMbench.
  }
  \label{fig:comparision_sft}
  \vspace{-6pt}
\end{figure}

{
\setlength{\tabcolsep}{1.mm}
\begin{table*}[t]
\centering
\footnotesize
\label{tab:exps_finetune}
\caption{
\footnotesize
\textbf{Illustration of the benefits of adding our \instructname data in the LVLM supervised fine-tuning (SFT) stage.}
We report the performance on our \benchmarkname and existing representative benchmarks including MMB (MMBench-Dev-EN~\cite{liu2023mmbench}), MMMU (MMMU-Val~\cite{yue2023mmmu}), MMStar~\cite{chen2024we}, MathVista~\cite{lu2024mathvista}, AI2D~\cite{kembhavi2016diagram}, HallBench~(HallusionBench~\cite{liu2023hallusionbench}), MMVet~\cite{yu2023mm} and ChartQA~\cite{masry2022chartqa}.
The best and second-best results in each section are colored \colorbox{firstBest}{\bf Green} and \colorbox{secondBest}{Red}, respectively.
}
\scalebox{.95}{
    \begin{tabular}{l| ccccccccc |c}
    \toprule
    \multirow{2}{*}{Method} & \multirow{2}{*}{\benchmarkname} & \multirow{2}{*}{MMB} & \multirow{2}{*}{MMMU} & MM & Math & \multirow{2}{*}{AI2D} & Hall & \multirow{2}{*}{MMVet} & Chart & \multirow{2}{*}{Avg.} \\
    ~ & &  & & Star & Vista & &Bench & & QA & \\
    \cmidrule(r){1-1} \cmidrule(r){2-10} \cmidrule(r){11-11}
    LLaVa1.5~\cite{liu2023improved} & \colorbox{secondBest}{32.2} & \colorbox{secondBest}{66.5}&\colorbox{secondBest}{35.7}&\colorbox{secondBest}{33.1}&\colorbox{secondBest}{25.2}&\colorbox{secondBest}{55.5}&\colorbox{firstBest}{\textbf{48.8}}&\colorbox{secondBest}{31.6}&\colorbox{secondBest}{21.2}&\colorbox{secondBest}{38.9}\\  
    LLaVa1.5 + \instructname & \colorbox{firstBest}{\textbf{37.2}} & \colorbox{firstBest}{\textbf{66.5}}&\colorbox{firstBest}{\textbf{37.4}}&\colorbox{firstBest}{\textbf{34.1}}&\colorbox{firstBest}{\textbf{25.2}}&\colorbox{firstBest}{\textbf{56.2}}&\colorbox{secondBest}{48.7}&\colorbox{firstBest}{\textbf{31.9}}&\colorbox{firstBest}{\textbf{23.4}}&\colorbox{firstBest}{\textbf{40.1}}\\   
    $\Delta$ & \hgreen{+5.0} & \hgreen{+0.0} & \hgreen{+1.7}& \hgreen{+1.0}& \hgreen{+0.0}& \hgreen{+0.7}& \hblue{-0.1}& \hgreen{+0.3}&\hgreen{+2.2} &\hgreen{+1.2}\\
    \cmidrule(r){1-1} \cmidrule(r){2-10} \cmidrule(r){11-11}
    InternLM-XC2~\cite{dong2024internlm} & \colorbox{secondBest}{35.6} & \colorbox{secondBest}{79.5} & \colorbox{secondBest}{41.4}&\colorbox{secondBest}{56.2}&\colorbox{secondBest}{57.2}&\colorbox{secondBest}{81.2}&\colorbox{secondBest}{60.0}&\colorbox{secondBest}{37.6}&\colorbox{secondBest}{62.6}&\colorbox{secondBest}{56.8}\\
    InternLM-XC2 + \instructname & \colorbox{firstBest}{\textbf{50.1}} & \colorbox{firstBest}{\textbf{79.9}}& \colorbox{firstBest}{\textbf{41.9}}&\colorbox{firstBest}{\textbf{57.3}}&\colorbox{firstBest}{\textbf{58.7}}&\colorbox{firstBest}{\textbf{81.2}}&\colorbox{firstBest}{\textbf{60.4}}&\colorbox{firstBest}{\textbf{38.8}}&\colorbox{firstBest}{\textbf{63.8}}&\colorbox{firstBest}{\textbf{59.1}}\\
    $\Delta$ & \hgreen{+14.5} & \hgreen{+0.4}&\hgreen{+0.5}& \hgreen{+1.1}& \hgreen{+1.5}& \hgreen{+0.0}&\hgreen{+0.4}&\hgreen{+1.2}&\hgreen{+1.2}&\hgreen{+2.3}\\
    \bottomrule
    \end{tabular}
}
\end{table*}
}
We showcase the superior quality of \instructname by presenting comparative results at the bottom of Tab.~\ref{tab:exps_mmdu}, where we incorporate \instructname into the SFT stage of LVLMs such as LLaVA 1.5~\cite{liu2023improved} and InternLM-XC2~\cite{dong2024internlm}. Results demonstrate that adding \instructname increases the overall performance on \benchmarkname, especially for the image relationship understanding ability. In Tab.~\ref{tab:exps_finetune}, we further demonstrate that integrating \instructname also benefits existing benchmarks that require multi-image understanding, such as MMMU~\cite{yue2023mmmu} and MMStar~\cite{chen2024we}, as well as short-form QA datasets like MMVet~\cite{yu2023mm}. To explain the performance improvement, we provide qualitative examples in Fig.~\ref{fig:comparision_sft}, illustrating that incorporating \instructname enables LVLMs to engage in longer and more accurate dialogues.

\subsection{Multi-Image Benchmark Results}\label{sec:multi}

Additionally, we test the model finetuned with \instructname on several multi-image benchmarks. We evaluate five benchmarks: MMMU~\cite{yue2023mmmu}, BLINK~\cite{fu2024blink}, Qbench2~\cite{wu2024qbench}, Mantis~\cite{jiang2024mantis}, and MMDU. For MMMU, only the results of multi-image questions are considered, and for Mantis, tests are conducted using both the "merge" and "sequence" methods. As shown in Tab.~\ref{tab:multi}, LLaVa1.5+\instructname achieved significant improvements across all multi-image benchmarks, with the most notable improvement observed in Mantis (sequence), reaching a $6.9\%$ increase. This indicates that \instructname greatly aids in enhancing the model's multi-image understanding capabilities, significantly addressing the model's shortcomings in reasoning within multi-image scenarios due to a lack of multi-image pre-train data.

\begin{table}[t]
\begin{minipage}{.68\textwidth}
\begin{center}
\setlength{\tabcolsep}{1pt}
\scalebox{0.80}{
\setlength{\tabcolsep}{3pt}
\begin{tabular}{cccc llllll}
\toprule
    \multirow{2}{*}{Models} & \multirow{2}{*}{MMMU} & \multirow{2}{*}{BLINK}  & \multirow{2}{*}{Qbench2}  & Mantis & Mantis & \multirow{2}{*}{MMDU} \\
    & (multi-pics) &  & & (sequence) & (merge) & \\
    \midrule
    LLaVa-1.5 & \colorbox{secondBest}{27.7} & \colorbox{secondBest}{37.1}  &  \colorbox{secondBest}{46.0} &  \colorbox{secondBest}{37.8} & \colorbox{secondBest}{41.9} &  \colorbox{secondBest}{32.2}  \\
    +\instructname & \colorbox{firstBest}{\textbf{29.8}} &   \colorbox{firstBest}{\textbf{40.1}} &   \colorbox{firstBest}{\textbf{48.5}} &  \colorbox{firstBest}{\textbf{44.7}} &  \colorbox{firstBest}{\textbf{44.7}}  &   \colorbox{firstBest}{\textbf{37.2}} \\
    $\Delta$  & \hgreen{+2.1} &  \hgreen{+3.0} &  \hgreen{+2.5} &  \hgreen{+6.9}  & \hgreen{+2.8}   & \hgreen{+5.0} \\
\bottomrule
\end{tabular}
}
\end{center}
\end{minipage}
\begin{minipage}{.3\textwidth}
\caption{
\small
\textbf{Multi-image benchmark test results.} We evaluated five multi-image benchmarks: MMMU~\cite{yue2023mmmu}, BLINK~\cite{fu2024blink}, Qbench2~\cite{wu2024qbench}, Mantis~\cite{jiang2024mantis}, and MMDU.
}
\label{tab:multi}
\end{minipage}
\vspace{-12pt}
\end{table}

\subsection{Ablation Study on Token Length and SFT Strategies}\label{sec:multi}
We test the LLaVa1.5 baseline (context-window length is 2-4k) and LLaVa1.5 (SFT on \instructname, we extend the context-window length to 8k with the RoPE interpolation) model with different context lengths. From Tab.~\ref{tab:ablation}, we observe that: (1) As the context length of the model increases, the performance also improves. (2) Finetuning on MMDU can increase the context window size of the LLaVA model.

Additionally, we compare different SFT strategies for training the model, including "Continue training" and "Add to the existing pool." The results in Tab.~\ref{tab:ablation} show that the final outcomes achieved by both methods are essentially the same.

\begin{table*}[!t]
    \centering
    \caption{\textbf{Ablation Study} on Token Length and SFT Strategies}
    \scalebox{.9}{
    \begin{tabular}{l|c|ccccccc}
    \toprule
        \multirow{2}{*}{Models} & Max & \multirow{2}{*}{C} & \multirow{2}{*}{R} & \multirow{2}{*}{VP} & \multirow{2}{*}{LC} & \multirow{2}{*}{AA} & \multirow{2}{*}{IRU} & Overall \\
        & tokens & & & & & & & Score \\
        \midrule
         \multirow{2}{*}{LLaVa-1.5} & 2k&  19.0&  19.0& 21.8&  29.3 &  22.5 &  19.6 &  20.9  \\
         ~ & 4k & 25.4 &  25.6 &  31.1 &  40.8 &  32.9 &  29.5 &  30.0\\
        \midrule
        \multirow{3}{*}{LLaVa-1.5+\instructname} & 2k & 20.0 & 20.1 &  22.1 &  29.4 &  23.5 &  21.6& 22.3\\
         ~ & 4k & \colorbox{secondBest}{31.5} & \colorbox{secondBest}{32.3} & \colorbox{secondBest}{34.9} & \colorbox{secondBest}{45.0} & \colorbox{secondBest}{36.3} & \colorbox{secondBest}{33.8} & \colorbox{secondBest}{34.9}\\
         ~ & 8k &\colorbox{firstBest}{\textbf{34.2}} & \colorbox{firstBest}{\textbf{34.3}} & \colorbox{firstBest}{\textbf{36.1}} & \colorbox{firstBest}{\textbf{48.2}} & \colorbox{firstBest}{\textbf{39.6}} & \colorbox{firstBest}{\textbf{34.7}} & \colorbox{firstBest}{\textbf{37.1}}\\
        \midrule
        Continue training & -  & 34.3& 34.5& 36.7& 47.2& 38.5& 35.5& 37.2 \\
        Add to the existing pool & -  & 34.3 & 36.3 & 37.1 & 47.3 & 38.9 & 35.7 & 37.3\\
    \bottomrule
    \end{tabular}}
    \label{tab:ablation}
    \vspace{-12pt}
\end{table*}
\section{Related Work}


\paragraph{LVLM Evaluation Benchmarks}
The rapid advancements in Large Vision-Language Models (LVLMs)\cite{bai2023qwen,claude3,openai2023gpt4,gpt4o,li2023monkey,laurençon2024matters,liu2023improved,lu2024deepseek,yu2024rlaifv,xu2024llava-uhd,young2024yi,chen2024far,dong2024internlm,bai2023qwen,liu2024llavanext,liu2024rar, 4khd, ma2024mmlongbench, liu2024mia} have spurred the development of comprehensive evaluation benchmarks to assess their capabilities across various tasks and domains. Numerous benchmarks~\cite{liu2024visual,liu2023mmbench,yin2023survey,li2023seed,li2023seed2,xu2023lvlm,ying2024mmt,yu2023mm,chen2024we} aim to provide a standardized and objective way to measure the performance of LVLMs and track their progress toward achieving general multi-modal understanding and reasoning.

Recently, specialized benchmarks have emerged to evaluate specific abilities~\cite{lu2022learn,liu2023hallusionbench}, such as for science reasoning~\cite{yue2023mmmu}, math reasoning~\cite{lu2024mathvista}, OCR recognition~\cite{liu2023hidden}, and diagram analysis~\cite{kembhavi2016diagram}.
Some existing benchmarks require multi-turn~\cite{liu2024convbench} chatting with a maximum of three turns, and others on multi-image comparison~\cite{suhr2018corpus,wu2024towards} with a maximum of four images. However, none of the existing benchmarks combine the multi-turn and multi-image abilities with a long context window for conversation applications, highlighting the need for a more comprehensive evaluation framework.

\paragraph{LVLM Instruction-Tuning Datasets}
The development of instruction tuning datasets for LLMs (\eg, Alpaca~\cite{alpaca}, Vicuna~\cite{vicuna2023}) has been instrumental in enhancing the instruction-following capabilities. Building upon the successes achieved in LLMs, researchers have proposed visual instruction tuning datasets (\eg, LLaVA-Instruct-150K~\cite{liu2024visual}, LLaVA 1.5-665K~\cite{liu2023improved}) to improve the instruction-following abilities of LVLMs. Moreover, several instruction-tuning datasets have been designed to enhance specific skills~\cite{dai2024instructblip,liu2023aligning,zhao2023svit}, such as ShareGPT4V~\cite{chen2023sharegpt4v} for caption generation, mPLUG-DocOwl~\cite{ye2023mplug} for document understanding, and VideoChatGPT~\cite{maaz2023video} for video comprehension. To the best of our knowledge, our \instructname is the first open-source multi-turn, multi-image, and long-context instruction tuning, making it a valuable resource for advancing human-AI interaction capabilities.
\section{Conclusion}\label{sec:conclusion}
\vspace{-6pt}

In this paper, we introduce \textbf{\benchmarkname}, a multi-mage, multi-turn, and long-context benchmark designed to enhance the daily human-AI conversation experience. Our comprehensive evaluation of 15 LVLMs reveals a significant performance disparity, with closed-source models like GPT-4o~\cite{gpt4o} outperforming the open-source LVLMs. This disparity may be attributed to the scarcity of open-source instruction tuning datasets that adequately assess the required multi-turn and multi-image abilities. To address this limitation and contribute to the open-source community, we propose \textbf{\instructname}, an instruction tuning dataset comprising 45k examples with a maximum of 17K text tokens, 5 images, and 9 turns. We also demonstrate that fine-tuning LVLMs on \instructname improves performance across various LVLM benchmarks. Our \textbf{\benchmarkname} and \textbf{\instructname} are poised to benefit the research community and foster future advancements in human-AI interaction.

\vspace{-6pt}
\paragraph{Limitations}\label{limitation} While \benchmarkname offers several advantages, we acknowledge two key limitations. (1) \benchmarkname primarily focuses on English and does not encompass multilingual abilities. (2) Our benchmark is designed to assess LVLMs' proficiency in daily scenarios, rather than specialized domain expertise (\eg, mathematical problem-solving in MathVista~\cite{lu2024mathvista}). By acknowledging these limitations, we hope to encourage future research directions and expansions of our benchmark such as incorporating multilingual support for other linguistic populations.

\vspace{-6pt}
\paragraph{Societal Impacts}\label{impacts} As \instructname is built upon Wikipedia, models fine-turned on \instructname may perpetuate biases and linguistic preferences in English. Moreover, LVLMs fine-tuned on our \instructname may be susceptible to factuality and hallucination issues, potentially generating inaccurate or misleading information. By recognizing these risks, we can work towards creating more inclusive, accurate, and reliable LVLMs that foster trustworthy human-AI interactions.

\vspace{-6pt}
\paragraph{Author Statement and Data License}
The authors bear all responsibility in case of violation of rights and confirm that this dataset is open-sourced under the \href{https://creativecommons.org/licenses/by-nc/4.0/deed.en}{Attribution-NonCommercial 4.0 International (CC BY-NC 4.0)} license. Using this dataset should abide by the \href{https://openai.com/policies/terms-of-use}{policy of OpenAI}.
\section*{Acknowledgments}
This project is funded in part by Shanghai Artificial lntelligence Laboratory, the National Key R$\&$D Program of China (2022ZD0160201), the Centre for Perceptual and Interactive Intelligence (CPII) Ltd under the Innovation and Technology Commission (ITC)’s InnoHK. Dahua Lin is a PI of CPII under the InnoHK.

{\small
\bibliographystyle{unsrt}
\bibliography{main}
}

\newpage
\appendix

\section*{\centering Appendices}
In this appendix, we offer further details regarding the proposed \benchmarkname and \instructname, along with additional experimental discussions aimed at comprehensive benchmarking. Specifically, Appendix \ref{appendix:open-source} includes our project URL and benchmark download URL. Appendix \ref{appendix:benchmark} and \ref{appendix:instruct} delve into the specifics of MMDU and MMDU-45k, respectively. Our evaluation Appendix \ref{appendix:evaluation} provides in-depth analysis and discussion.
Appendix \ref{appendix:datasheet} provides the datasheets for \benchmarkname and \instructname.

\section{Open-source Links}\label{appendix:open-source}

All data from our \benchmarkname and \instructname are now available for viewing or download via the following URLs:
\begin{itemize}
    \item Project page: https://liuziyu77.github.io/MMDU/
    \item GitHub repository: https://github.com/Liuziyu77/MMDU
    \item \benchmarkname benchmark: https://huggingface.co/datasets/laolao77/MMDU
    \item \instructname instruction tuning dataset: https://huggingface.co/datasets/laolao77/MMDU
    \item URL to Croissant metadata: https://huggingface.co/datasets/laolao77/MMDU
\end{itemize}

\section{More Details of \benchmarkname}\label{appendix:benchmark}
We present the details of our \benchmarkname, encompassing the pipeline of our data cluster, the prompt designed for dialogue generation, visualizations of our generated examples, and a comprehensive comparison between our \benchmarkname and existing benchmarks.

\subsection{Data preparation}
\begin{figure}[t]
  \centering
  \includegraphics[width=.95\linewidth]{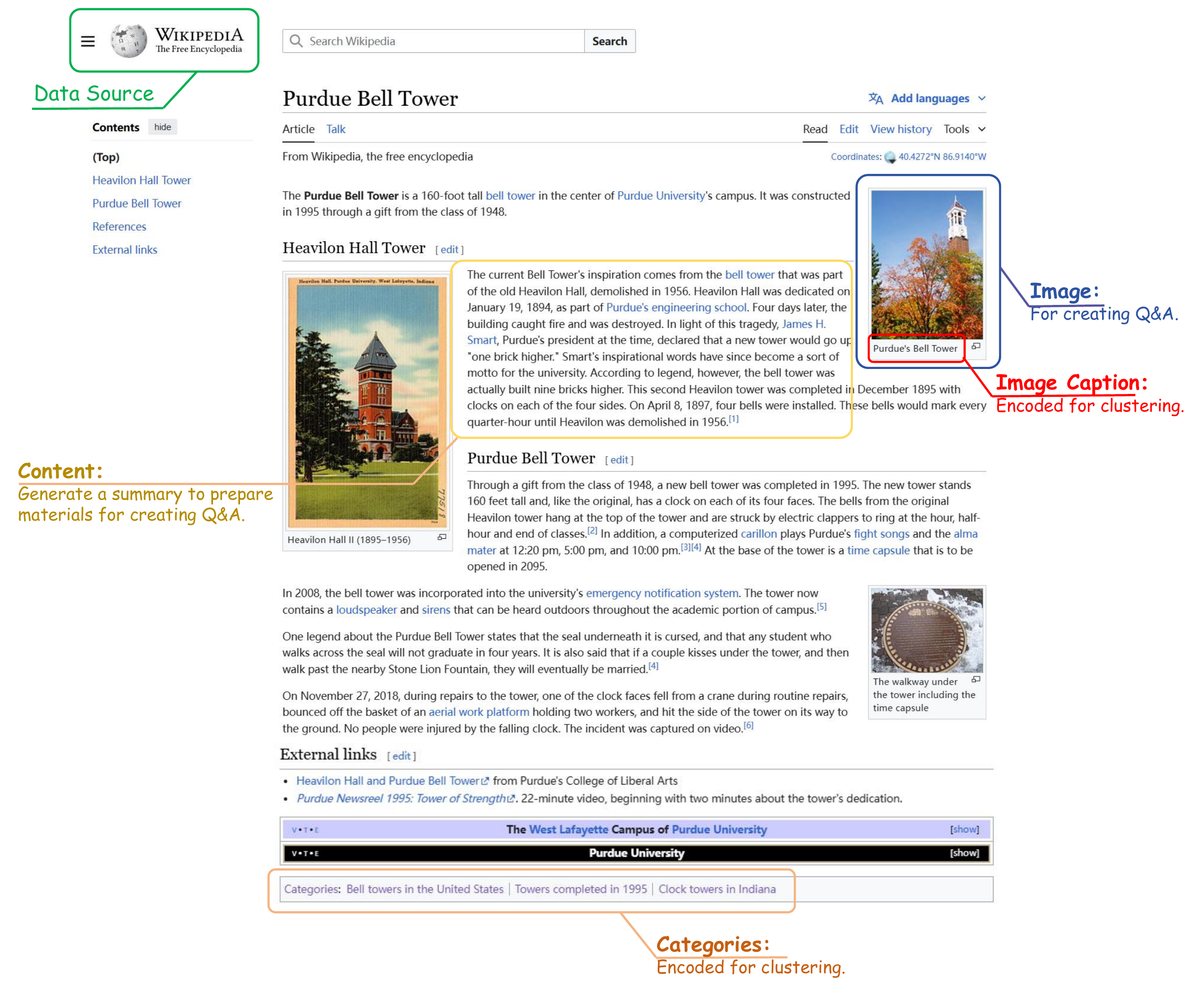}
  \vspace{-6pt}
  \caption{\textbf{Usage of Wikipedia information}. We primarily use Wikipedia's images, captions, content, and categories.
  }
  \label{fig:wikicase}
\end{figure}

In this section, we provide a detailed explanation of how to use data from Wikipedia~\cite{srinivasan2021wit} to construct \benchmarkname. As shown in Fig. \ref{fig:wikicase}, for a Wikipedia entry, we first obtain the entry's images, image captions, main content, and categories. The primary function of captions and categories (tags) is to cluster the entries. The captions, main content of the entries, and images are mainly used to generate multi-image, multi-round dialogues.

\begin{figure}[t]
  \centering
  \includegraphics[width=.95\linewidth]{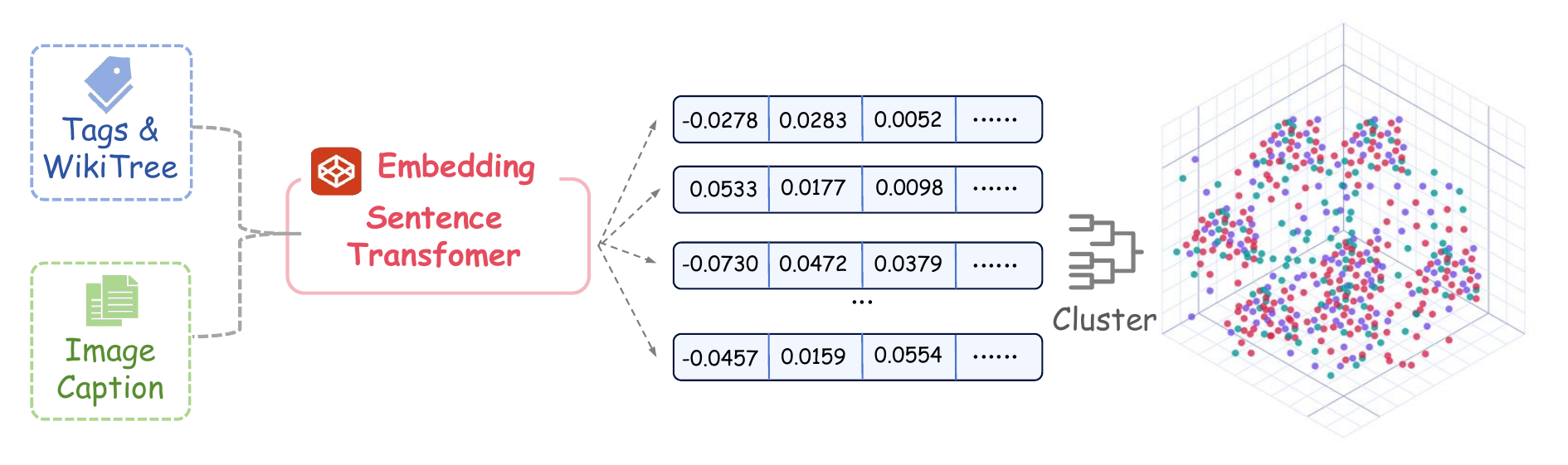}
  \vspace{-6pt}
  \caption{\textbf{Clustering pipeline.} We use clustering methods to process Wikipedia entries, grouping together entries with high relevance.
  }
  \label{fig:cluster}
\end{figure}

Fig. \ref{fig:cluster} illustrates how we use captions and categories (tags) for clustering the entries. Subsequently, as shown in Fig. 2 in the main text, the main content of the entry is processed by InternLM-chat-20B\cite{cai2024internlm2} to generate a summary of the entry. This summary, along with the image captions and images, are then input into GPT-4o to generate multi-image, multi-round dialogue content.

\begin{figure}[t]
  \centering
  \includegraphics[width=.95\linewidth]{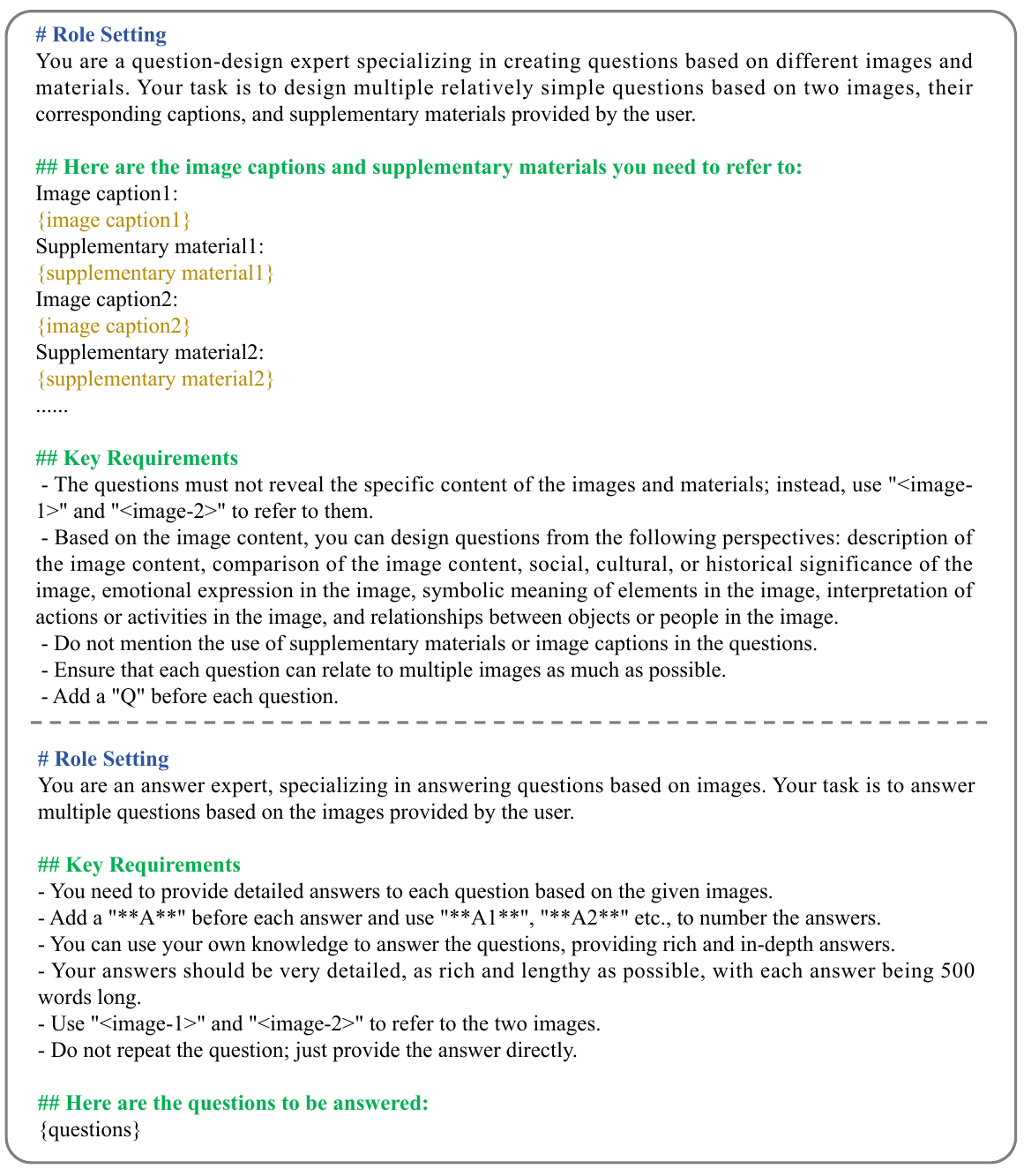}
  \vspace{-6pt}
  \caption{\textbf{Dialogue generation prompt}. Here, we present a prompt for generating a dialogue based on two images.
  }
  \label{fig:creatqa}
\end{figure}

\begin{figure}[t]
  \centering
  \includegraphics[width=1.0\linewidth]{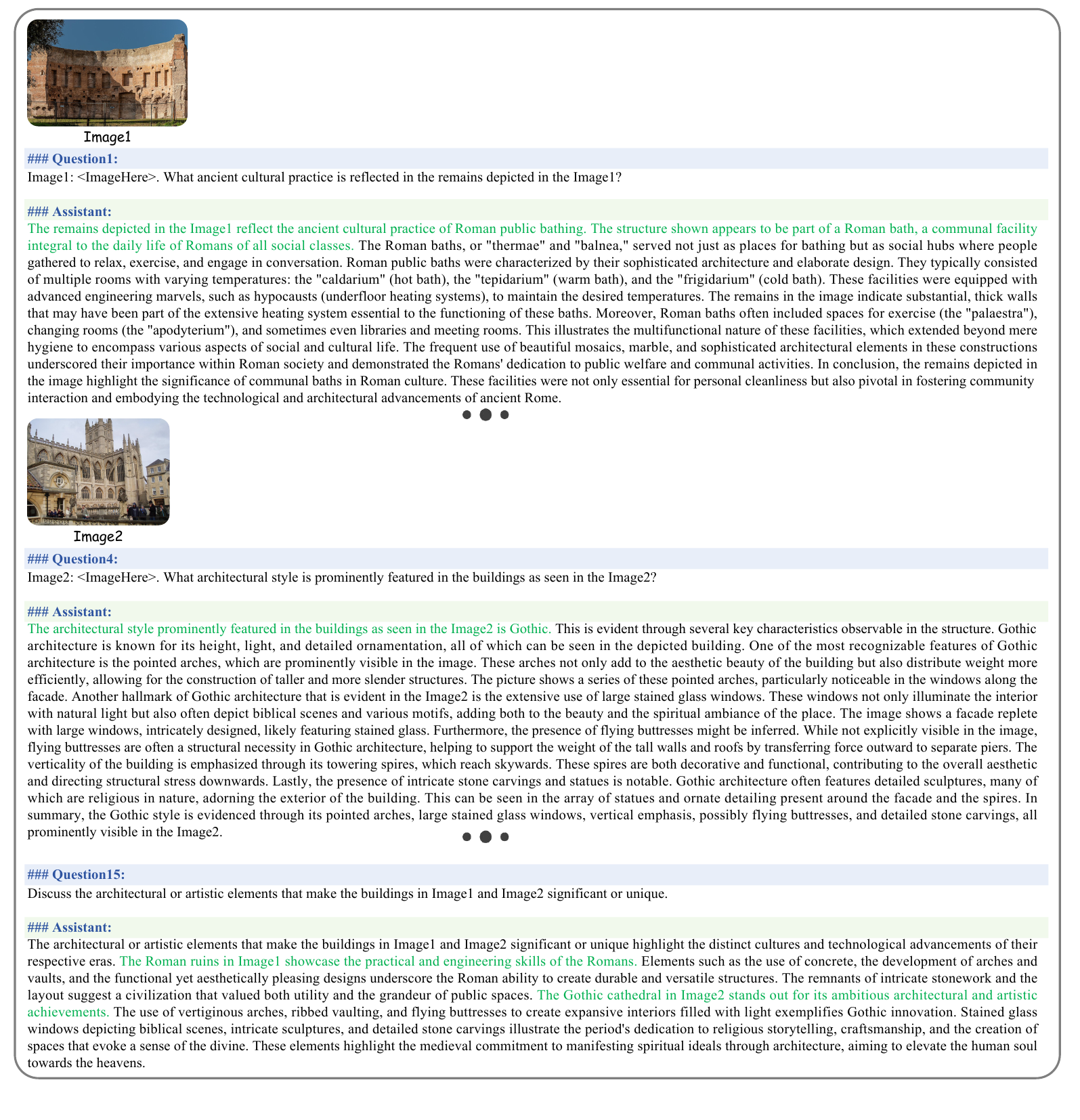}
  \vspace{-12pt}
  \caption{\textbf{Benchmark Example (a)}. Here, we present a 2-image benchmark multi-round Q\&A task. Due to space limitations, we only display the three rounds of dialogue.
  }
  \label{fig:archcase}
\end{figure}

\subsection{Prompt of Dialogue generation}
To use the images and content clustered in Fig. \ref{fig:cluster} effectively, we design a prompt, illustrated in Fig. \ref{fig:creatqa}, to assist GPT-4o in crafting multi-turn questions. Firstly, our prompt incorporates both the image and its accompanying content, facilitating GPT-4o to pose insightful and pertinent questions pertaining to the image's theme. We employ GPT-4o to generate a multitude of questions centered around the image theme, drawing from both the images and their textual context. Secondly, to ensure the wide usability of the data, we avoid providing textual cues when generating answers. Instead, we task GPT-4o with comprehending and addressing multiple questions solely based on the images themselves and their interrelations. This approach yields multi-turn questions and answers that evince a profound grasp of the images and are applicable across various contexts.

Specifically, in Fig. \ref{fig:creatqa}, the upper part illustrates the prompt for question generation. In the ``Role Setting'' segment, we instruct GPT-4o to assume the role of a ``question-design expert'', tasked with formulating questions inspired by a variety of images and materials. To foster depth and breadth, we delineate the content parameters in the ``Key Requirements'' segment, encompassing descriptors, comparisons, social and cultural contexts, historical significance, emotional nuances, symbolic interpretations, and relational inquiries.

The prompt for generating answers is depicted in the lower part of Fig. \ref{fig:creatqa}. Similarly guided by the ``Role Setting'' segment, GPT-4o is required to serve as an ``answer expert'' and respond to the generated questions based solely on the images. As there is no reliance on highly specific textual knowledge, the content of GPT-4o's answers will tend to be more generalized. Additionally, we employ ``<image-i>'' in both the generated questions and answers to denote the position of the image, allowing for the rearrangement of image positions by substituting ``i''. This theoretically allows for the generation of conversations that include more images and longer sequences of questions and answers.

\subsection{Example Visualization of \benchmarkname}
In this section, we illustrate several examples to qualitatively assess the quality of our \benchmarkname. Example (a) of Fig. \ref{fig:archcase} illustrates numerous turns of questions and answers. Question 1 and Question 4 pertain to the content of images 1 and 2, respectively, while Question 15 revisits the content of both images, challenging LVLMs' long-text comprehension and memory capabilities. 

\begin{figure}[t]
  \centering
  \includegraphics[width=1.0\linewidth]{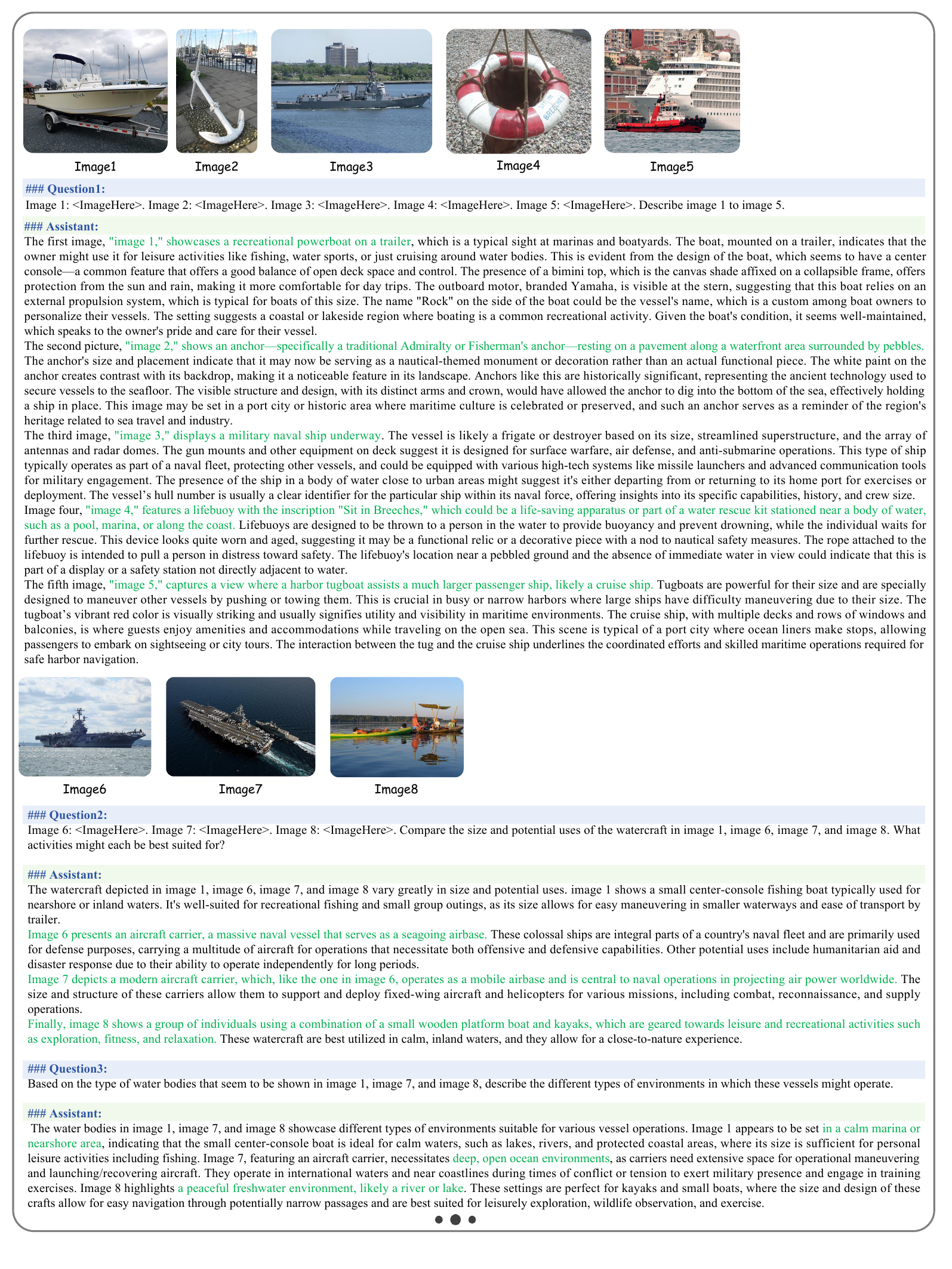}
  \vspace{-12pt}
  \caption{\textbf{Benchmark Example (b)}. Here, we present a multi-image benchmark multi-round Q\&A task. Due to space limitations, we only display the first three rounds of dialogue.
  }
  \vspace{-12pt}
  \label{fig:shipcase1}
\end{figure}

As depicted in Example (b) of Fig. \ref{fig:shipcase1}, Question 1 entails analyzing five images. We begin by identifying the position of each image using the placeholder ``<image-i>'' and subsequently insert the corresponding images marked as ``<ImageHere>''. The task in Question 1 entails providing detailed descriptions for each image. For instance, image 1 depicts a ``recreational powerboat on a trailer'', image 2 shows a ``traditional Admiralty or Fisherman's anchor'', image 3 features a ``military naval ship underway'', image 4 displays a ``lifebuoy'', and image 5 portrays a ``view where a harbor tugboat assists a much larger passenger ship''. Question 2 and Question 3 expand the inquiry with 3 additional images each,  employing the same way to introduce them. Question 2 delves into the practical applications of the subjects within the images, while Question 3 explores their applicable environments. Both questions elicit detailed descriptions in the responses. These questions, along with their comprehensive answers, serve as valuable references for evaluating LVLMs' comprehension of image content and their interrelations.

\begin{figure}[t]
  \centering
  \includegraphics[width=1.0\linewidth]{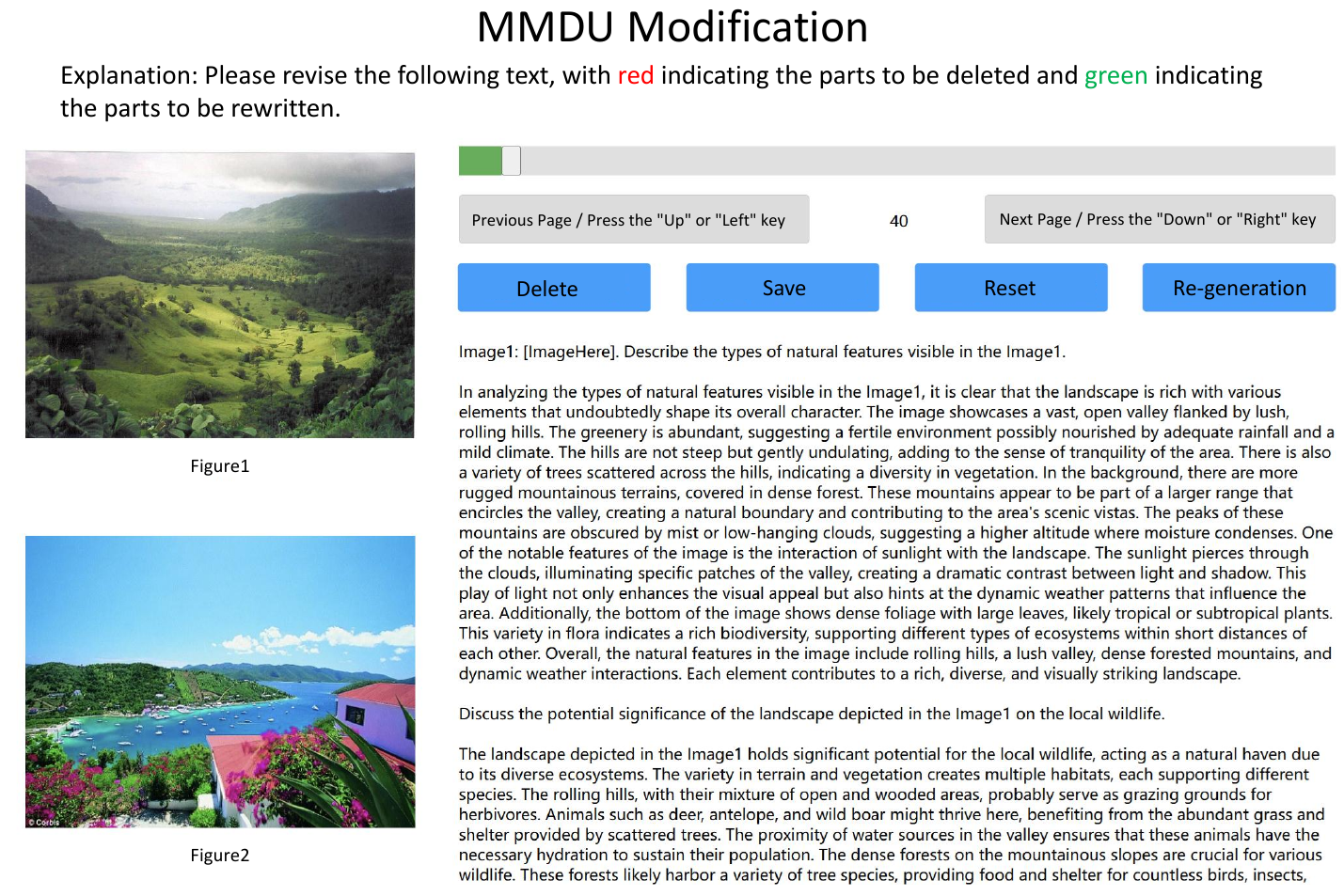}
  \vspace{-12pt}
  \caption{\textbf{Web UI} for human annotators.
  }
  \vspace{-12pt}
  \label{fig:webui}
\end{figure}

\begin{figure}[t]
  \centering
  \includegraphics[width=.9\linewidth]{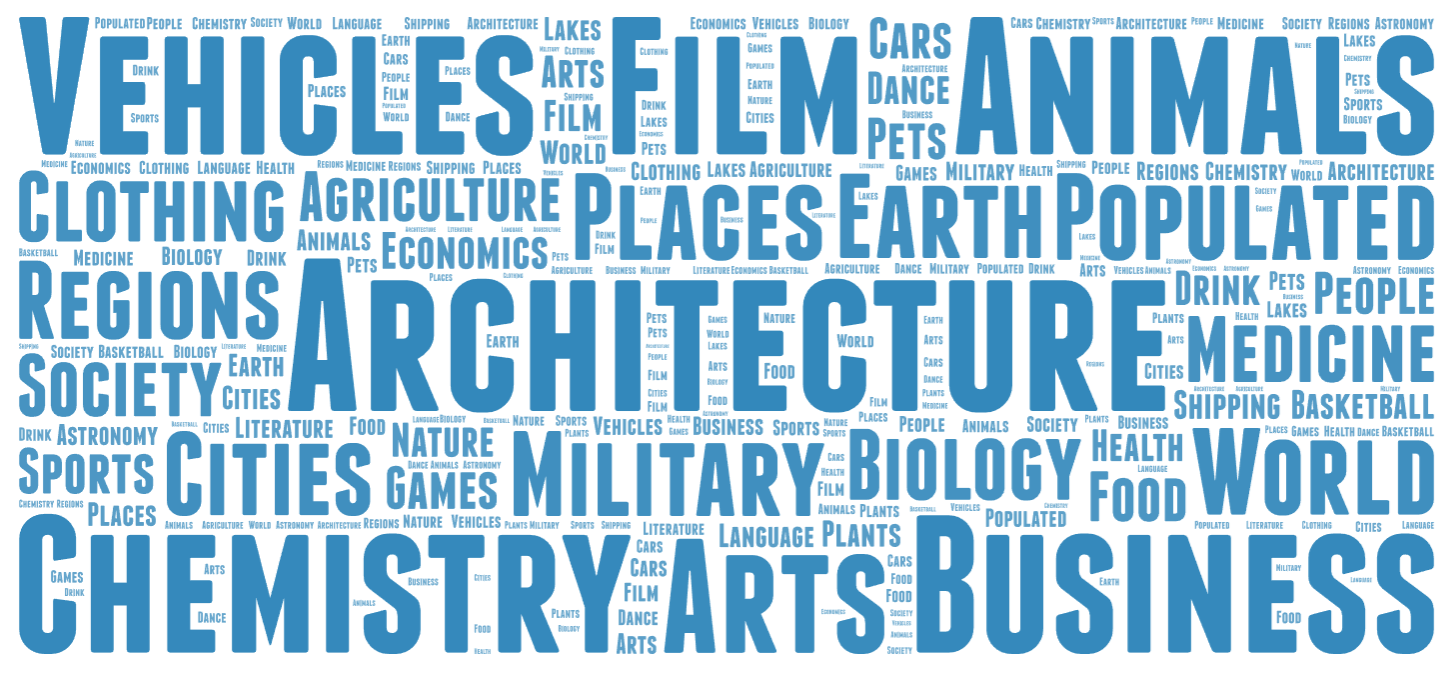}
  \vspace{-12pt}
  \caption{\textbf{Category word cloud of \instructname}.
  }
  \vspace{-12pt}
  \label{fig:ciyun}
\end{figure}

\subsection{Related Work with Existing Benchmarks}
We conduct discussion with existing benchmarks, including MMVet~\cite{yu2023mm}, MMBench~\cite{liu2023mmbench}, MMMU~\cite{yue2023mmmu}, MMStar~\cite{chen2024we}, MathVista~\cite{lu2024mathvista}, AI2D~\cite{kembhavi2016diagram}, HallusionBench~\cite{liu2023hallusionbench}, Chart QA~\cite{masry2022chartqa} and ConvBench~\cite{liu2024convbench}. 

\textbf{MMVet} is a benchmark designed to evaluate LVLMs’ ability on complicated multimodal tasks. It contains  200 images, and 218 questions with the corresponding answers, each set pertaining to one image. Our \benchmarkname offers 421 images with 1645 Q\&A pairs and around 15 turns of each Q\&A pair for 2-20 images.

\textbf{MMBench} contains over 3000 multiple-choice questions covering 20 different ability dimensions, such as object localization and social reasoning, for evaluating LVLMs. However, multiple-choice questions fail to adequately assess the generative and conversational capabilities of LVLMs.

\textbf{MMMU} includes 11.5K meticulously collected multimodal questions from college exams, quizzes, and textbooks, covering six core disciplines. These questions span 30 subjects and 183 subfields, comprising 30 highly heterogeneous image types, such as charts, diagrams, maps, tables, music sheets, and chemical structures. 

\textbf{MMStar} is an elite vision-indispensable multi-modal benchmark comprising 1,500 samples meticulously selected by humans. MMStar benchmarks 6 core capabilities and 18 detailed axes, aiming to evaluate LVLMs’ multi-modal capacities with carefully balanced and purified samples.

\textbf{MathVista} is a benchmark designed to combine challenges from diverse mathematical and visual tasks. It consists of 6,141 examples, derived from 28 existing multimodal datasets involving mathematics and 3 newly created datasets. 

\textbf{AI2D} is a dataset of diagrams with exhaustive annotations of constituents and relationships for over 5,000 diagrams and 15,000 questions and answers. It is designed to evaluate LVLMs' ability to interpret and reason about intricate diagrams with meticulous attention to detail and clarity.

\textbf{HallusionBench} is a comprehensive benchmark designed for the evaluation of image-context reasoning, which comprises 346 images paired with 1129 questions. HallusionBench primarily tests the issues of language hallucination and visual illusion present in LVLMs.

\textbf{Chart QA} is a large-scale benchmark covering 9.6K human-written questions as well as 23.1K questions generated from human-written chart summaries. It focuses on assessing LVLMs' abilities with charts.

\textbf{ConvBench} evaluates multi-turn conversations by assessing perception, reasoning, and creativity progressively. It comprises 577 multi-turn conversations aligned with their respective single images. 

\subsection{More Details about Human Annotators}\label{appendix:webui}
We designed a dedicated data manipulation Web UI for manual data inspection, and its interface is shown in Fig.~\ref{fig:webui}.

\section{More Details of \instructname}\label{appendix:instruct}
In this section, we provide more detailed information about \instructname. The data construction method for \instructname is essentially the same as that of the \benchmarkname. This section mainly introduces some of the notable features of \instructname.

\subsection{The powerful scalability of \instructname}
The powerful scalability of the  \instructname dataset can be attributed to the well-designed data format we implemented during its construction. For all generated data, we use the identifier "<image-i>" to mark the positions and sequences of all images. For data generated in different batches, we can stack different multi-image, multi-round dialogues by modifying the sequence identifier "i" in "<image-i>". This allows us to construct dialogues with longer contexts and more images according to user requirements for dialogue length.

 \instructname acts like a fundamental building block, enabling users to construct dialogues of any desired length without having to collect images and textual information from scratch. Instead, users can use  \instructname as a component to build training data or test questions tailored to their specific needs.

\subsection{The Richness and Diversity of \instructname}
During the construction of \instructname, we performed clustering on data from Wikipedia. In the clustering process, all Wikipedia entries were categorized into various groups. As shown in Fig. \ref{fig:ciyun}, these categories include geography, history, culture, nature, animals, plants, vehicles, mathematics, physics, chemistry, and more. This distribution ensures that the MMDU-45k dataset has a very broad coverage, encompassing various aspects of daily life. Consequently, using \instructname for training allows the model to learn more general knowledge and enhances its capabilities in long dialogues and multi-image understanding across multiple domains.

\section{More Details}\label{appendix:evaluation}
In this section, we present a comprehensive overview of our evaluation details, including specific judgment prompts, as well as quantitative and qualitative results.

\subsection{Judgment prompt}
\begin{figure}[h]
  \centering
  \includegraphics[width=.9\linewidth]{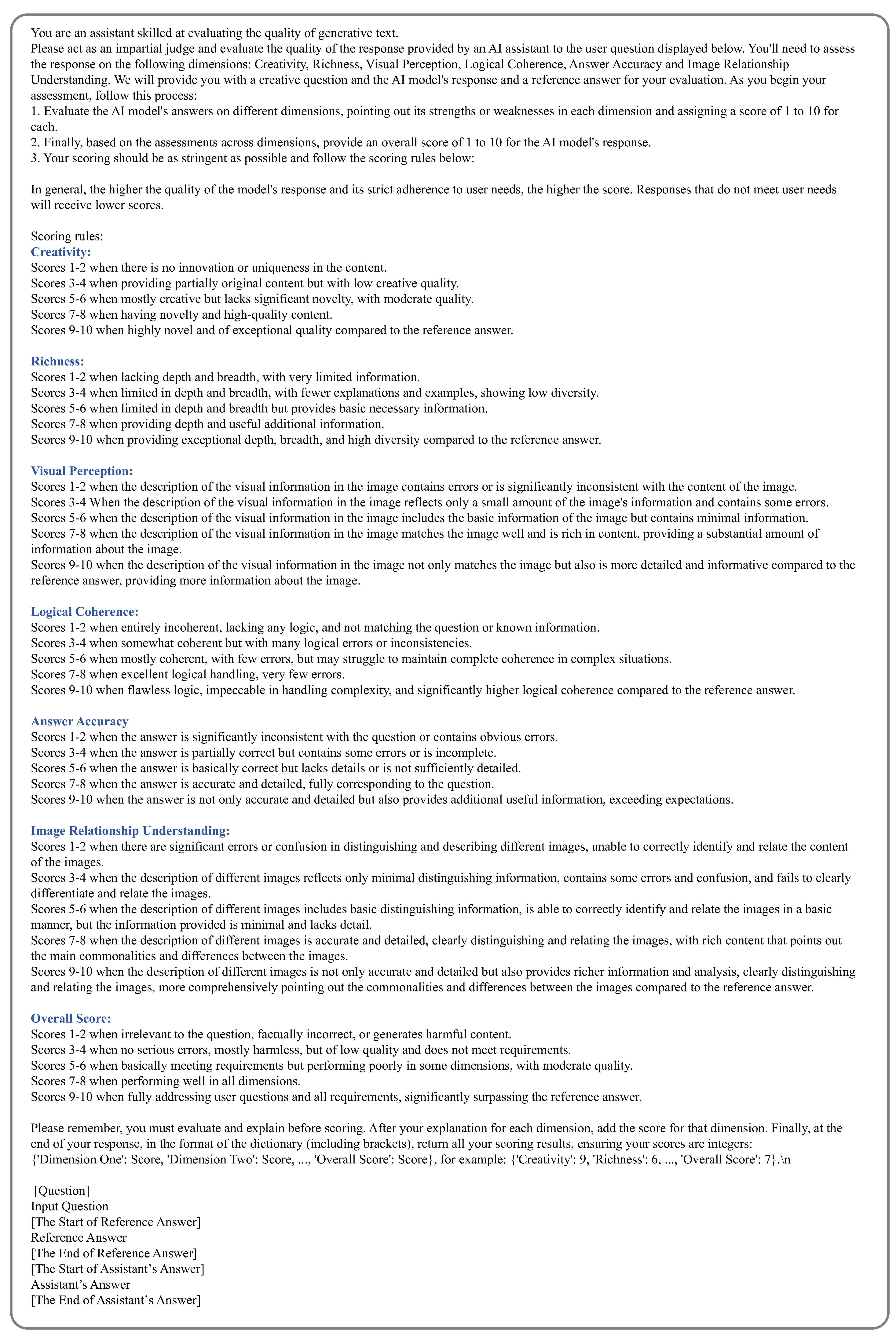}
  \vspace{-12pt}
  \caption{\textbf{Judgment prompt} used to test the results of GPT-4, GPT-4-Turbo, and Claude3-Opus.
  }
  \label{fig:judgeprompt}
\end{figure}

In Fig. \ref{fig:judgeprompt}, we illustrate the judgment prompt employed to guide GPT-4o in conducting comprehensive evaluations of LVLM results. This process involves delineating evaluation criteria across six dimensions: Creativity, Richness, Visual Perception, Logical Coherence, Answer Accuracy, and Image Relationship Understanding. Each dimension is finely scored on a scale of 0 to 10, with criteria set at 2-point intervals, and supported by reference answers. Furthermore, GPT-4o is tasked with assigning an Overall Score, also at 2-point intervals. Finally, we divide the total score by the number of questions and multiply by 10 to obtain the final result.

Through this meticulous guidance, GPT-4o can effectively evaluate LVLM results across various dimensions, providing a comprehensive assessment process to validate the soundness of its scoring.

\begin{table*}[t]
    \footnotesize
    \begin{center}
    \caption{
    \footnotesize
    \textbf{Evaluate the results of the four models using different judgment models.} We used GPT-4o, GPT-4-turbo, and Claude3-Opus to evaluate the results of the four models.
    }
    \label{tab:otherpoint}
    \renewcommand{\arraystretch}{0.5}
    \scalebox{.9}{
    \begin{tabular}{{@{}l | l| c c c c c c |c  @{}}}
    \toprule
    Models & Judgment Models & C & R & VP & LC & AA & IRU & Avg.\\
    \midrule
    \multirow{3}{*}{LLaVa1.5-7B~\cite{liu2023improved}} & GPT-4o & 27.8 & 28.0 & 33.2 & 43.0 & 35.4 & 31.7 & 32.2\\
    & GPT-4-turbo & 28.2 & 27.9 & 34.2 & 39.9 & 34.8 & 32.3 & 32.1\\
    & Claude3-opus &  32.8 & 32.6 & 40.6 & 47.1 & 41.1 & 29.0 & 39.3 \\ 
    \midrule
    \multirow{3}{*}{LLaVa1.5-7B+\instructname} & GPT-4o & 34.3 & 34.5 & 36.7 & 47.2 & 38.5 & 35.5 & 37.2\\
    & GPT-4-turbo & 36.2 & 37.4 & 39.3 & 47.4 & 40.8 & 38.3 & 39.1\\
    & Claude3-opus & 43.0 & 42.3 & 51.0 & 57.0 & 51.8 & 37.6 & 49.3 \\ 
    \midrule
    \multirow{3}{*}{Claude3-Opus~\cite{claude3}} & GPT-4o & 58.6 & 61.5 & 59.7 & 75.1 & 64.1 & 59.8 & 62.6 \\
    & GPT-4-turbo & 59.9 & 64.9 & 63.7 & 73.5 & 66.2 & 63.1 & 64.5 \\
    & Claude3-opus & 64.7 & 67.3 & 74.5 & 80.1 & 76.8 & 60.6 & 72.7 \\
    \midrule
    \multirow{3}{*}{GPT-4-turbo~\cite{openai2023gpt4}} & GPT-4o & 62.0 & 64.2 & 63.4 & 78.0 & 69.0 & 64.4 & 66.3\\
    & GPT-4-turbo & 63.9 & 67.6 & 67.7 & 76.1 & 70.8 & 67.0 & 68.4\\
    & Claude3-opus & 65.9 & 67.9 & 74.5 & 80.5 & 76.9 & 60.9 & 73.7 \\ 
    \midrule
    \multirow{3}{*}{GPT-4o~\cite{gpt4o}} & GPT-4o & 63.7 & 69.6 & 66.7 & 80.6 & 73.3 & 68.1 & 70.2\\
    & GPT-4-turbo &  64.9 & 70.7 & 68.7 & 77.2 & 73.2 &68.6  & 70.1\\
    & Claude3-opus & 67.5 & 71.9 & 76.2 & 82.3 & 79.4 & 64.1 & 75.9\\ 
    \bottomrule     
    \end{tabular}}
    \vspace{-12pt}
    \end{center}
\end{table*}

\begin{figure}[t]
  \centering
  \includegraphics[width=.9\linewidth]{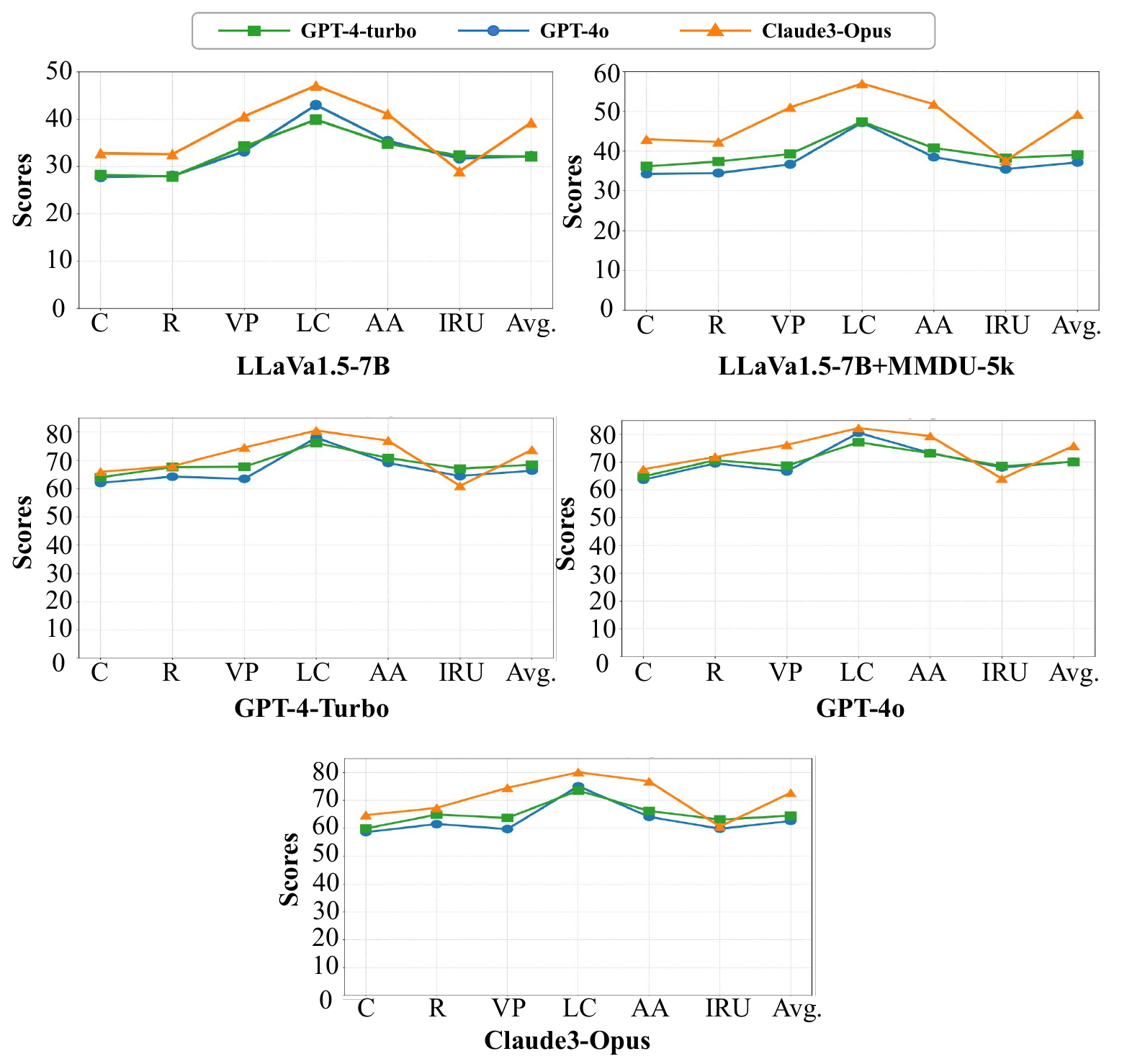}
  \vspace{-12pt}
  \caption{\textbf{Judgment Results.} We used GPT-4o, GPT-4-Turbo, and Claude3-Opus as judgment models to test the performance of LLaVa1.5-7B, LLaVa1.5-7B+MMDU-45k, GPT-4o, GPT-4-Turbo, and Claude3-Opus on MMDU.}
  \label{fig:zexian}
\end{figure}

\textbf{Different Judgement Models.} We conduct a comparative analysis of evaluation using GPT-4o, GPT-4-turbo and Claude3-Opus across various LVLMs, presented in Tab. \ref{tab:otherpoint} and Fig. \ref{fig:zexian}. From the results in the table, we can observe that the scoring trends of GPT-4o and GPT-4-turbo are similar, with minimal differences. The scores provided by the Claude3-Opus model show a similar trend to those of GPT-4o and GPT-4-turbo but are generally slightly higher. Additionally, for the IRU (Image Relationship Understanding) metric, the scores given by Claude3-Opus are more conservative compared to the other two models, being slightly lower than those of GPT-4o and GPT-4-turbo. However, the findings overall show a strong similarity between the evaluation outcomes of GPT-4, GPT-4 Turbo, and Claude3-Opus, highlighting the robustness of our proposed judgment prompt and evaluation pipeline.

\textbf{Consistency with Human Scoring.} Furthermore, we quantify the concordance between the scoring of GPT-4-turbo and Claude3-Opus with human judgment. In contrast to GPT-4o, which exhibits Pearson, Spearman, and Kendall similarities of 97.5\%, 97.3\%, and 89.0\% respectively, Claude3-Opus and GPT-4-turbo demonstrates Pearson, Spearman, and Kendall similarities of (91.4\%, 92.7\%, 89.0\%) and (97.2\%, 97.0\%, 88.5\%), respectively. These metrics indicate that while Claude and GPT-4-turbo closely align with human scores, its performance slightly trails behind the more potent GPT-4o.

\subsection{More cases}
\begin{figure}[h]
  \centering
  \includegraphics[width=.91\linewidth]{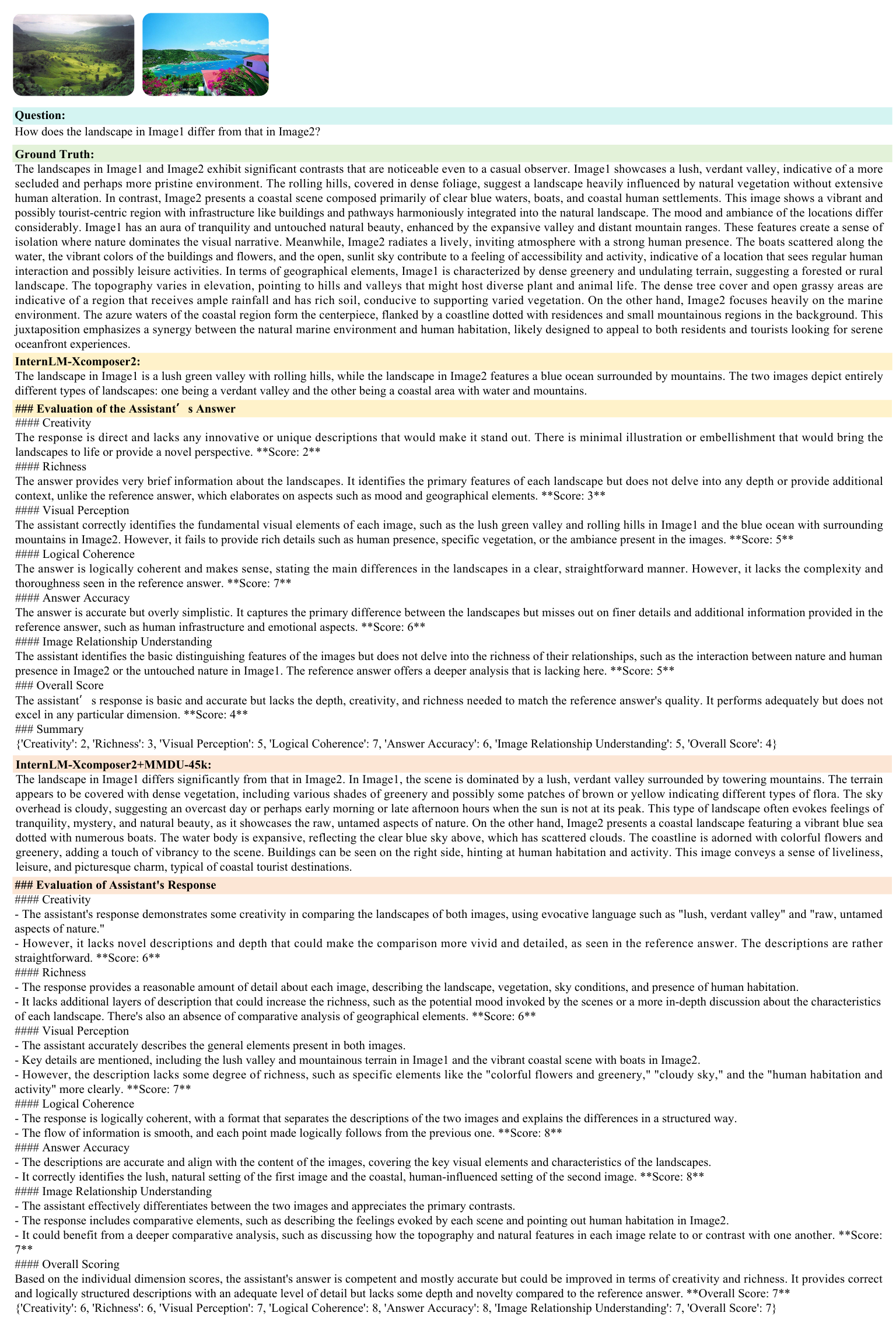}
  \vspace{-12pt}
  \caption{\textbf{Q\&A Case 1 of MMDU.} We list the questions, images, and ground truth. We compared the results of InternLM-XC2 and InternLM-XC2+\instructname and showcase the scoring results and reasons for GPT-4o.
  }
  \vspace{-12pt}
  \label{fig:morecase1}
\end{figure}

\begin{figure}[h]
  \centering
  \includegraphics[width=.93\linewidth]{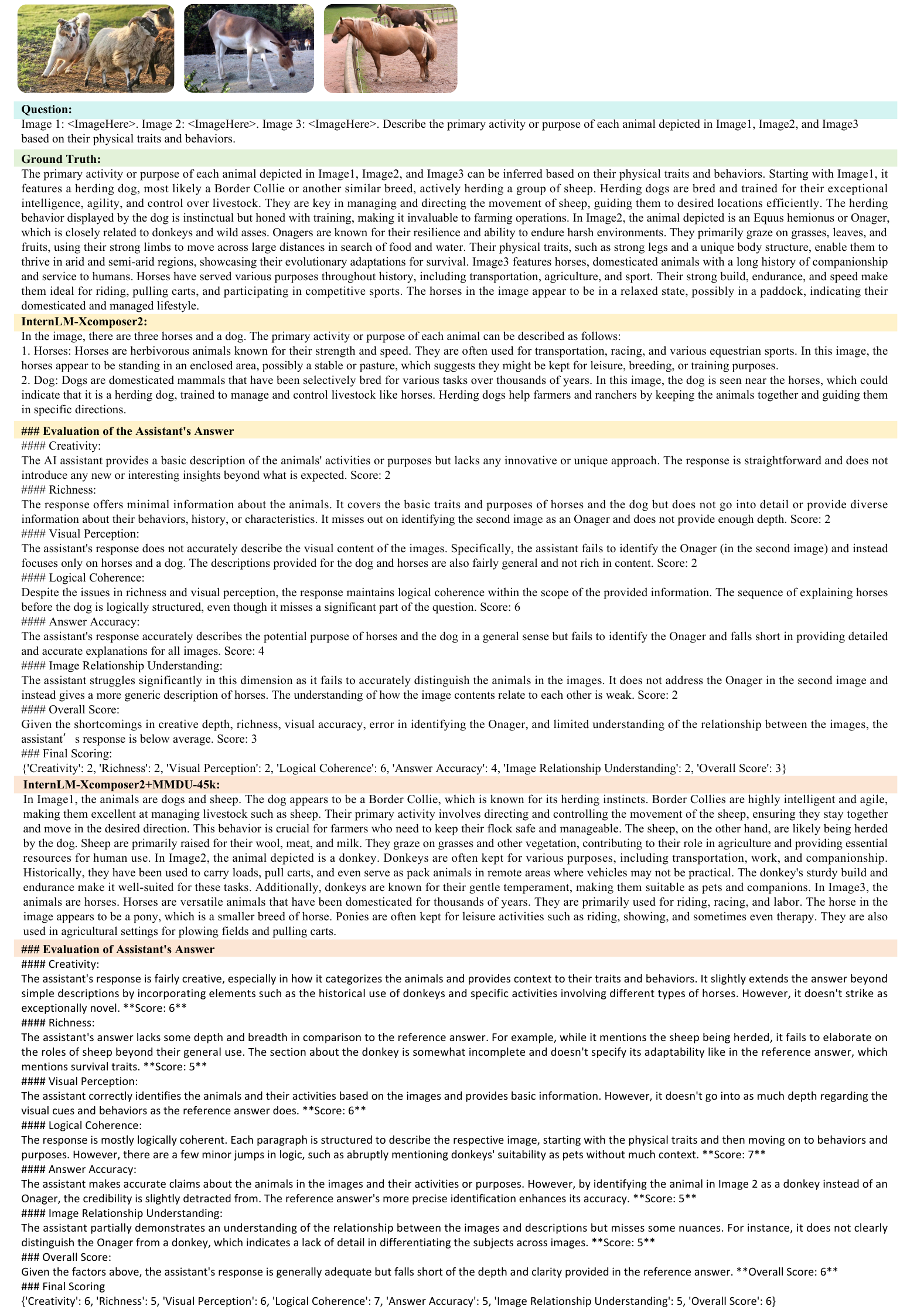}
  \vspace{-12pt}
  \caption{\textbf{Q\&A Case 2 of MMDU.}
  }
  \vspace{-12pt}
  \label{fig:morecase2}
\end{figure}

\begin{figure}[h]
  \centering
  \includegraphics[width=.93\linewidth]{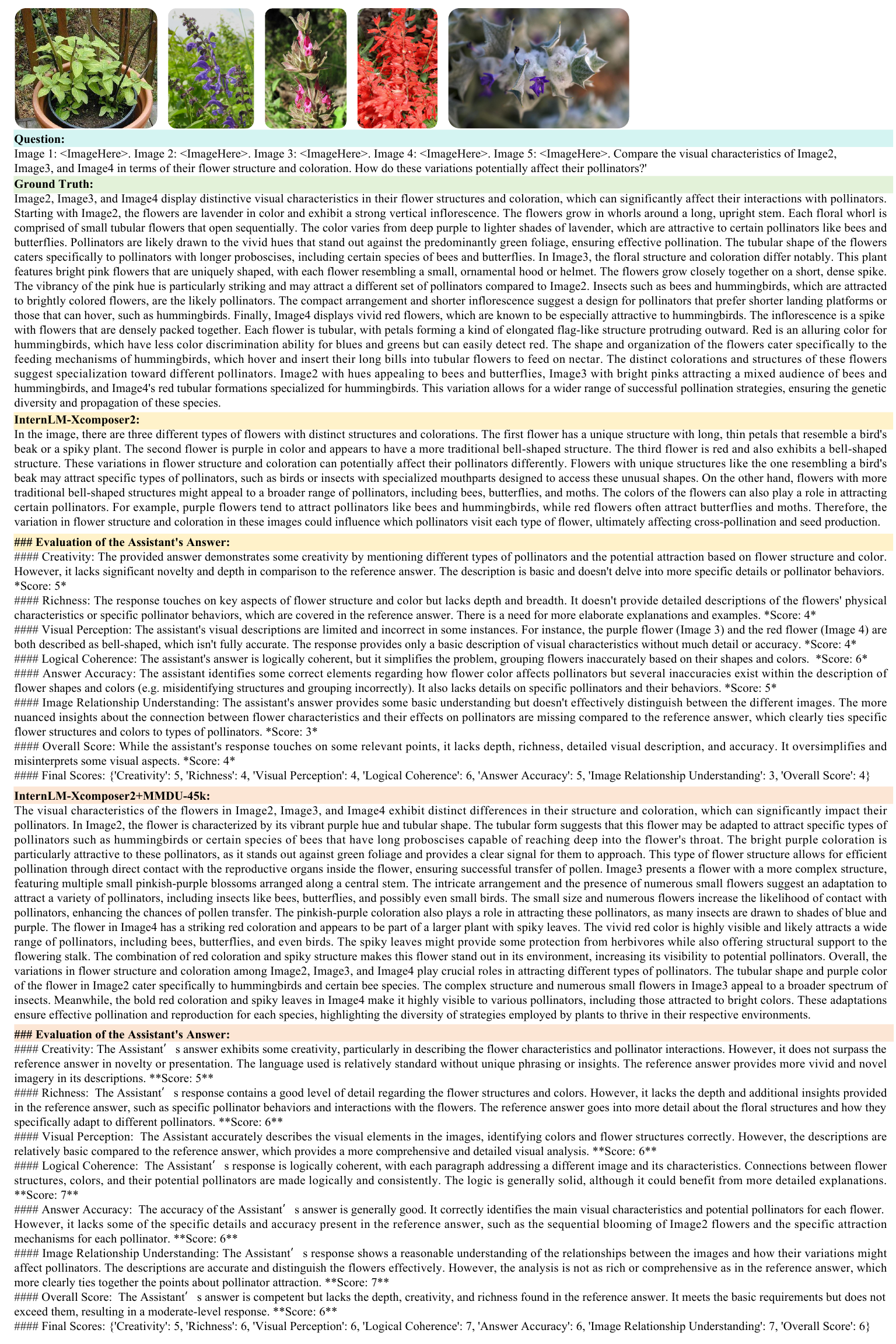}
  \vspace{-12pt}
  \caption{\textbf{Q\&A Case 3 of MMDU.}
  }
  \label{fig:morecase3}
  \vspace{-12pt}
\end{figure}

To clarify the testing and evaluation process of MMDU, we display three question-answer pairs from \benchmarkname in Fig. \ref{fig:morecase1}, \ref{fig:morecase2} and \ref{fig:morecase3}. Due to space limitations, we cannot show a complete multi-turn, multi-image conversation, so we selected one question-answer pair for demonstration. Each case includes the relevant images and the ground truth. We also list the results of InternLM-Xcomposer2 and InternLM-Xcomposer2+\instructname, showcasing the effectiveness of \instructname in improving the model's ability to handle multi-image, multi-turn conversations. Additionally, we provide the scoring results using GPT-4o, including the reasons and given scores.

\subsection{Finetune details}
In the experimental section, we finetuned the LLaVa1.5-7B and InternLM-Xcomposer2 using our \instructname dataset. During the finetuning process, we mixed the llava-665k dataset with the MMDU-45k dataset. Our learning rate was set to 2e-4, and we ran the training for 1 epoch.

\subsection{Cluster Accuracy}

In the process of constructing multi-item clusters, we ensure clustering accuracy through two key steps: We first utilize the inherent tags or labels associated with each wiki item for clustering. These tags and labels are manually annotated in wiki items, providing a high level of accuracy. Additionally, we further employ image captions for clustering, ensuring that the resulting clusters are largely free of noise and errors.

\begin{figure}[h]
  \centering
  \includegraphics[width=1.\linewidth]{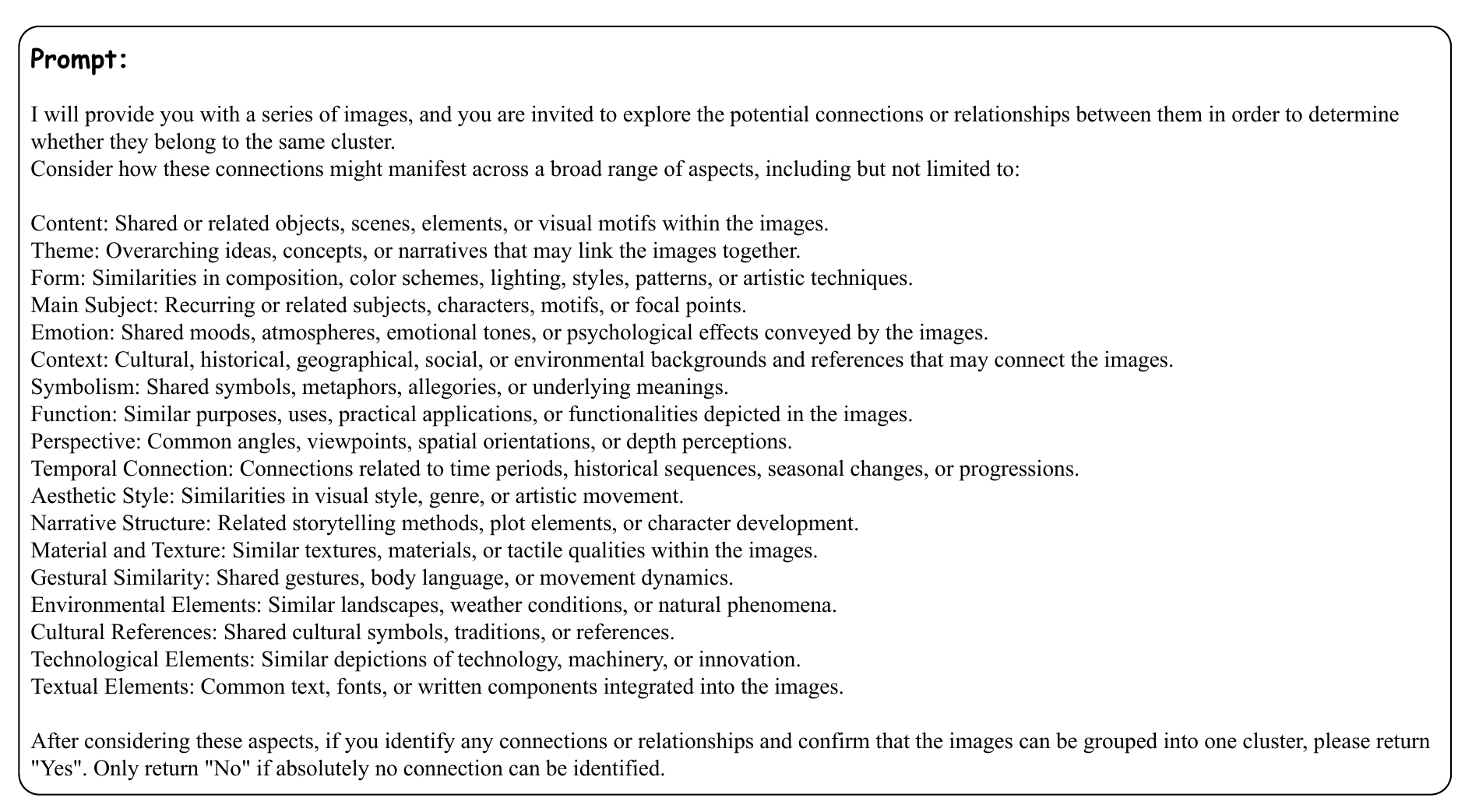}
  \vspace{-12pt}
  \caption{\textbf{Prompt for Clustering Accuracy Verification}
  }
  \label{fig:cluster_accuracy_prompt}
  \vspace{-12pt}
\end{figure}

To verify the accuracy of the clustering, we conducted the following experiment: We randomly sampled five sets of data from the \instructname dataset, each containing 100 entries. The images from each entry were input into the GPT-4o model, which was tasked with evaluating whether the multiple images were related and could be grouped into the same cluster. The resulting accuracy rates were recorded as follows: $94\%$, $90\%$, $92\%$, $89\%$, $91\%$. This indicates that the clustering accuracy of \benchmarkname is very high. The prompt is shown in Fig.~\ref{fig:cluster_accuracy_prompt}.

\subsection{Reasoning in \benchmarkname}

When answering questions from \benchmarkname, the model must process interleaved image and text inputs while simultaneously understanding the relationships between multiple images which requires various visual reasoning capabilities. We believe that \benchmarkname's visual reasoning capabilities can be reflected in the following aspects:

\textbf{Visual Perception Reasoning:} to interpret and infer information from single or multiple images, recognizing relevant details, patterns, and relationships between the images.

\textbf{Visual Relationship Reasoning:} to understand and articulate the connections and relationships between different images, determining how they interact or contribute to the overall dialogue.

\textbf{High-Level Semantic Reasoning:} to grasp the deeper meanings, symbols, or abstract concepts represented across multiple images.

\textbf{Cross-modal reasoning:} to extract visual information from an image and combine it with textual or conversational cues to infer knowledge or information that goes beyond the content of the image.

\begin{figure*}[t]
    \centering
    \includegraphics[width=1.0\linewidth]{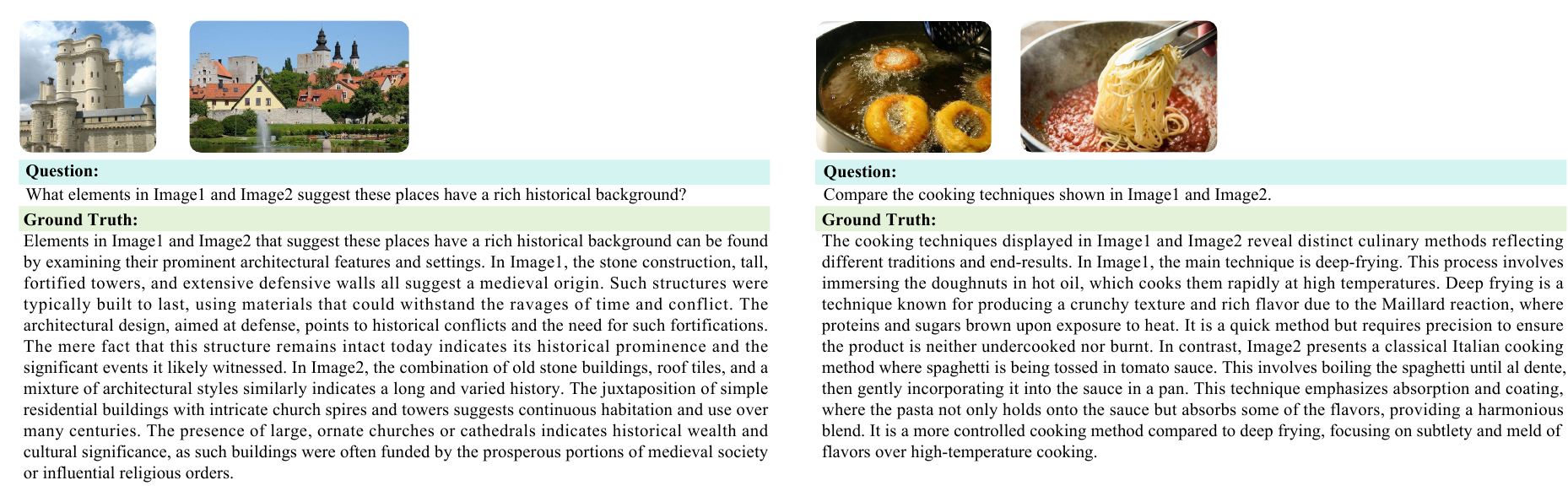}
    \vspace{-6pt}
    \caption{\textbf{Left: Visual Perception Reasoning.} The most basic visual reasoning. \textbf{Right: Visual Relationship Reasoning.} Understand and articulate the connections and relationships between different images.}
    \label{fig:reasoning1}
    \vspace{-6pt}
\end{figure*}

\begin{figure*}[h]
    \centering
    \includegraphics[width=1.0\linewidth]{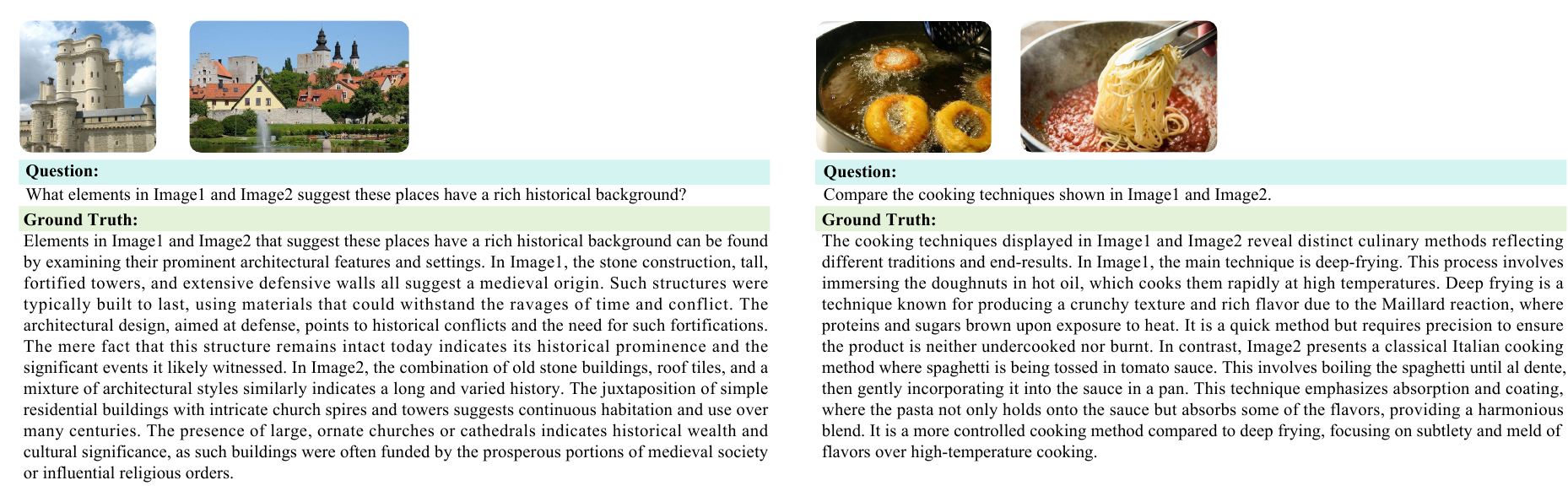}
    \vspace{-6pt}
    \caption{\textbf{Left: High-Level Semantic Reasoning.} Grasping the deeper meanings, symbols, or abstract concepts across multiple images. \textbf{Right: Cross-modal reasoning.} Extracting visual information from an image and combining it with textual cues to infer knowledge or information that goes beyond the content of the image.}
    \label{fig:reasoning2}
    \vspace{-6pt}
\end{figure*}

These reasoning abilities are also considered when constructing the QA pairs for \benchmarkname. In Fig.~\ref{fig:reasoning1} and Fig.~\ref{fig:reasoning2}, we've included cases that illustrate the dataset's visual reasoning capabilities.


\section{Datasheet for Datasets}\label{appendix:datasheet}
The following section contains answers to questions listed in datasheets for datasets.

\subsection{Motivation}

\begin{itemize}
  \item  For what purpose was the dataset created?\\ 
  \benchmarkname and \instructname are created to serve as a benchmark for evaluate and improve LVLM's abilities in multi-turn and multi-image conversations
  \item Who created the dataset (e.g., which team, research group) and on
behalf of which entity (e.g., company, institution, organization)?
  \\ The authors of this paper.
  \item Who funded the creation of the dataset? If there is an associated
grant, please provide the name of the grantor and the grant name and
number.
\\This work was supported by Shanghai AI Lab.

\end{itemize}

\subsection{Composition}

\begin{itemize}
  \item  What do the instances that comprise the dataset represent (e.g., documents, photos, people, countries)?
  \\ \benchmarkname and \instructname consist of multi-image, multi-turn dialogues. Each dialogue includes a dozen or so rounds of questions and answers, as well as multiple images.
  \item How many instances are there in total (of each type, if appropriate)?
  \\ \benchmarkname consists of 110 dialogues, comprising 1,645 questions and their corresponding answers. Additionally, these 110 dialogues in \benchmarkname contain 421 images. On the other hand, \instructname is composed of 45,000 dialogues, with each dialogue averaging 3 images and 9 rounds of questions and answers, totaling 410k Q\&A pairs.
  \item Does the dataset contain all possible instances or is it a sample (not necessarily random) of instances from a larger set?
  \\
\instructname is a new dataset generated using data from Wikipedia. \benchmarkname's data is partly a sample selected from \instructname and partly manually generated data.
  \item  What data does each instance consist of?
  \\ 
  Each instance contains multiple images and rounds of question-answer dialogues. The number of images and the number of Q\&A rounds per instance are not consistent. Each instance in \benchmarkname contains 2-20 images, while each instance in \instructname contains 2-5 images. On average, each instance in \benchmarkname contains 15 rounds of Q\&A, whereas each instance in \instructname contains an average of 9 rounds of Q\&A.
  \item  Is there a label or target associated with each instance?
  \\ Yes, in \benchmarkname, multiple questions within each dialogue have labels that have been manually checked and modified.
  \item  Is any information missing from individual instances? If so, please provide a description, explaining why this information is missing (e.g., because it was unavailable). This does not include intentionally removed information, but might include, e.g., redacted text.
  \\ N/A.
  \item  Are relationships between individual instances made explicit (e.g., users’ movie ratings, social network links)?
  \\ N/A.
  \item  Are there recommended data splits (e.g., training, development/validation, testing)?
  \\ Yes, \instructname is the training set, while \benchmarkname is the test set.
  \item  Are there any errors, sources of noise, or redundancies in the dataset?
  \\ N/A.
  \item  Is the dataset self-contained, or does it link to or otherwise rely on external resources (e.g., websites, tweets, other datasets)?
  \\ The dataset is self-contained.
  \item  Does the dataset contain data that might be considered confidential (e.g., data that is protected by legal privilege or by doctor– patient confidentiality, data that includes the content of individuals’ non-public communications)?
  \\ N/A.
  \item  Does the dataset contain data that, if viewed directly, might be offensive, insulting, threatening, or might otherwise cause anxiety?
  \\ N/A.
  \item Does the dataset relate to people?
  \\ Yes.
  \item  Does the dataset identify any subpopulations (e.g., by age, gender)? 
  \\ N/A.
  \item  Is it possible to identify individuals (i.e., one or more natural persons), either directly or indirectly (i.e., in combination with other data) from the dataset?
  \\ N/A.
  \item  Does the dataset contain data that might be considered sensitive in any way (e.g., data that reveals race or ethnic origins, sexual orientations, religious beliefs, political opinions or union memberships, or locations; financial or health data; biometric or genetic data; forms of government identification, such as social security numbers; criminal history)?
  \\ N/A.  
\end{itemize}

\subsection{Collection Process}

\begin{itemize}
  \item  How was the data associated with each instance acquired?
  \\ We used the open-source data from Wikipedia, incorporating its entries (including text and images), and applied GPT-4 to construct our own dataset.
  \item  What mechanisms or procedures were used to collect the data (e.g., hardware apparatuses or sensors, manual human curation, software programs, software APIs)?
  \\ We used entries collected from Wikipedia as our data source and then applied clustering methods to process the data.
  \item  If the dataset is a sample from a larger set, what was the sampling strategy (e.g., deterministic, probabilistic with specific sampling probabilities)?
  \\ N/A.
  \item   Who was involved in the data collection process (e.g., students, crowdworkers, contractors) and how were they compensated (e.g., how much were crowdworkers paid)?
  \\ The co-authors of the paper participated in the data collection, verification, and modification of \benchmarkname.
  \item  Over what timeframe was the data collected?
  \\ The data was collected in May of 2024, but the results do not depend much on the date of date collection.
  \item  Were any ethical review processes conducted (e.g., by an institutional review board)?
  \\ N/A.
  \item Does the dataset relate to people?
  \\ Yes.  
  \item  Did you collect the data from the individuals in question directly, or obtain it via third parties or other sources (e.g., websites)?
  \\ We obtained data from the open-source Wikipedia.
  \item  Were the individuals in question notified about the data collection?
  \\ We didn't collect the data from the individuals.
  \item  Did the individuals in question consent to the collection and use of their data?
  \\  We didn't collect the data from the individuals.
  \item  If consent was obtained, were the consenting individuals provided with a mechanism to revoke their consent in the future or for certain uses?
  \\ N/A.
  \item  Has an analysis of the potential impact of the dataset and its use on data subjects (e.g., a data protection impact analysis) been conducted?
  \\ The dataset does not have individual-specific information.
\end{itemize}

\subsection{Preprocessing/cleaning/labeling}

\begin{itemize}
  \item  Was any preprocessing/cleaning/labeling of the data done (e.g., discretization or bucketing, tokenization, part-of-speech tagging, SIFT feature extraction, removal of instances, processing of missing values)?
  \\ We performed clustering on the obtained data, selecting those with higher coherence. We also removed data with low-resolution images. Furthermore, we conducted manual checks and modifications on the \benchmarkname data to ensure its quality.
  \item  Was the “raw” data saved in addition to the preprocessed/cleaned/labeled data (e.g., to support unanticipated future uses)?
  \\ Yes.
  \item  Is the software that was used to preprocess/clean/label the data available?
  \\ Preprocessing, cleaning, and labeling are done via Python.
\end{itemize}

\subsection{Uses}

\begin{itemize}
  \item  Has the dataset been used for any tasks already?
  \\ No.
  \item  Is there a repository that links to any or all papers or systems that use the dataset?
  \\ No.
  \item  What (other) tasks could the dataset be used for?
  \\ N/A.
  \item  Is there anything about the composition of the dataset or the way it was collected and preprocessed/cleaned/labeled that might impact future uses?
  \\ N/A.
  \item  Are there tasks for which the dataset should not be used?
  \\ N/A.
\end{itemize}

\subsection{Distribution}

\begin{itemize}
  \item  Will the dataset be distributed to third parties outside of the entity (e.g., company, institution, organization) on behalf of which the dataset was created?
  \\ No.
  \item  How will the dataset will be distributed (e.g., tarball on website, API, GitHub)?
  \\ The dataset will be released on Huggingface.
  \item  When will the dataset be distributed?
  \\ The dataset will be released in mid-June 2024.
  \item  Will the dataset be distributed under a copyright or other intellectual property (IP) license, and/or under applicable terms of use (ToU)?
  \\ The dataset will be released under the \href{https://creativecommons.org/licenses/by-nc/4.0/deed.en}{Attribution-NonCommercial 4.0 International (CC BY-NC 4.0)} license.
  \item Have any third parties imposed IP-based or other restrictions on the data associated with the instances?
  \\ No.
  \item Do any export controls or other regulatory restrictions apply to the dataset or to individual instances?
  \\ No.
\end{itemize}

\subsection{Maintenance}

\begin{itemize}
  \item  Who will be supporting/hosting/maintaining the dataset?
  \\ The authors of this paper.
  \item  How can the owner/curator/manager of the dataset be contacted (e.g., email address)?
  \\ Contact the first author or other authors.
  \item  Is there an erratum?
  \\ No.
  \item  Will the dataset be updated (e.g., to correct labeling errors, add new instances, delete instances)?
  \\ If any correction is needed, we plan to upload a new version.
  \item If the dataset relates to people, are there applicable limits on the retention of the data associated with the instances (e.g., were the individuals in question told that their data would be retained for a fixed period of time and then deleted)? 
  \\ N/A
  \item Will older versions of the dataset continue to be supported/hosted/maintained?
  \\ Yes.
  \item If others want to extend/augment/build on/contribute to the dataset, is there a mechanism for them to do so?
  \\ Contact the authors of the paper.
\end{itemize}

\newpage
\section*{NeurIPS Paper Checklist}

\begin{enumerate}

\item {\bf Claims}
    \item[] Question: Do the main claims made in the abstract and introduction accurately reflect the paper's contributions and scope?
    \item[] Answer: \answerYes{} 
    \item[] Justification: The abstract and introduction clearly state our contributions, and the claims match the experimental results.  
    \item[] Guidelines:
    \begin{itemize}
        \item The answer NA means that the abstract and introduction do not include the claims made in the paper.
        \item The abstract and/or introduction should clearly state the claims made, including the contributions made in the paper and important assumptions and limitations. A No or NA answer to this question will not be perceived well by the reviewers. 
        \item The claims made should match theoretical and experimental results, and reflect how much the results can be expected to generalize to other settings. 
        \item It is fine to include aspirational goals as motivation as long as it is clear that these goals are not attained by the paper. 
    \end{itemize}

\item {\bf Limitations}
    \item[] Question: Does the paper discuss the limitations of the work performed by the authors?
    \item[] Answer: \answerYes{} 
    \item[] Justification: The limitations are discussed in Section~\ref{limitation}. 
    \item[] Guidelines:
    \begin{itemize}
        \item The answer NA means that the paper has no limitation while the answer No means that the paper has limitations, but those are not discussed in the paper. 
        \item The authors are encouraged to create a separate "Limitations" section in their paper.
        \item The paper should point out any strong assumptions and how robust the results are to violations of these assumptions (e.g., independence assumptions, noiseless settings, model well-specification, asymptotic approximations only holding locally). The authors should reflect on how these assumptions might be violated in practice and what the implications would be.
        \item The authors should reflect on the scope of the claims made, e.g., if the approach was only tested on a few datasets or with a few runs. In general, empirical results often depend on implicit assumptions, which should be articulated.
        \item The authors should reflect on the factors that influence the performance of the approach. For example, a facial recognition algorithm may perform poorly when image resolution is low or images are taken in low lighting. Or a speech-to-text system might not be used reliably to provide closed captions for online lectures because it fails to handle technical jargon.
        \item The authors should discuss the computational efficiency of the proposed algorithms and how they scale with dataset size.
        \item If applicable, the authors should discuss possible limitations of their approach to address problems of privacy and fairness.
        \item While the authors might fear that complete honesty about limitations might be used by reviewers as grounds for rejection, a worse outcome might be that reviewers discover limitations that aren't acknowledged in the paper. The authors should use their best judgment and recognize that individual actions in favor of transparency play an important role in developing norms that preserve the integrity of the community. Reviewers will be specifically instructed to not penalize honesty concerning limitations.
    \end{itemize}

\item {\bf Theory Assumptions and Proofs}
    \item[] Question: For each theoretical result, does the paper provide the full set of assumptions and a complete (and correct) proof?
    \item[] Answer: \answerNA{} 
    \item[] Justification: The paper does not include theoretical results. 
    \item[] Guidelines:
    \begin{itemize}
        \item The answer NA means that the paper does not include theoretical results. 
        \item All the theorems, formulas, and proofs in the paper should be numbered and cross-referenced.
        \item All assumptions should be clearly stated or referenced in the statement of any theorems.
        \item The proofs can either appear in the main paper or the supplemental material, but if they appear in the supplemental material, the authors are encouraged to provide a short proof sketch to provide intuition. 
        \item Inversely, any informal proof provided in the core of the paper should be complemented by formal proofs provided in appendix or supplemental material.
        \item Theorems and Lemmas that the proof relies upon should be properly referenced. 
    \end{itemize}

    \item {\bf Experimental Result Reproducibility}
    \item[] Question: Does the paper fully disclose all the information needed to reproduce the main experimental results of the paper to the extent that it affects the main claims and/or conclusions of the paper (regardless of whether the code and data are provided or not)?
    \item[] Answer: \answerYes{} 
    \item[] Justification:  Detailed instructions for replicating the results are provided in Appendix~\ref{appendix:open-source}. Additionally, the code, model checkpoint, and data will be publicly released. 
    \item[] Guidelines:
    \begin{itemize}
        \item The answer NA means that the paper does not include experiments.
        \item If the paper includes experiments, a No answer to this question will not be perceived well by the reviewers: Making the paper reproducible is important, regardless of whether the code and data are provided or not.
        \item If the contribution is a dataset and/or model, the authors should describe the steps taken to make their results reproducible or verifiable. 
        \item Depending on the contribution, reproducibility can be accomplished in various ways. For example, if the contribution is a novel architecture, describing the architecture fully might suffice, or if the contribution is a specific model and empirical evaluation, it may be necessary to either make it possible for others to replicate the model with the same dataset, or provide access to the model. In general. releasing code and data is often one good way to accomplish this, but reproducibility can also be provided via detailed instructions for how to replicate the results, access to a hosted model (e.g., in the case of a large language model), releasing of a model checkpoint, or other means that are appropriate to the research performed.
        \item While NeurIPS does not require releasing code, the conference does require all submissions to provide some reasonable avenue for reproducibility, which may depend on the nature of the contribution. For example
        \begin{enumerate}
            \item If the contribution is primarily a new algorithm, the paper should make it clear how to reproduce that algorithm.
            \item If the contribution is primarily a new model architecture, the paper should describe the architecture clearly and fully.
            \item If the contribution is a new model (e.g., a large language model), then there should either be a way to access this model for reproducing the results or a way to reproduce the model (e.g., with an open-source dataset or instructions for how to construct the dataset).
            \item We recognize that reproducibility may be tricky in some cases, in which case authors are welcome to describe the particular way they provide for reproducibility. In the case of closed-source models, it may be that access to the model is limited in some way (e.g., to registered users), but it should be possible for other researchers to have some path to reproducing or verifying the results.
        \end{enumerate}
    \end{itemize}

\item {\bf Open access to data and code}
    \item[] Question: Does the paper provide open access to the data and code, with sufficient instructions to faithfully reproduce the main experimental results, as described in supplemental material?
    \item[] Answer: \answerYes{} 
    \item[] Justification: The data source link is presented in Appendix~\ref{appendix:open-source}. Detailed information to reproduce all experimental results is provided in Appendix~\ref{appendix:evaluation}.
    \item[] Guidelines:
    \begin{itemize}
        \item The answer NA means that paper does not include experiments requiring code.
        \item Please see the NeurIPS code and data submission guidelines (\url{https://nips.cc/public/guides/CodeSubmissionPolicy}) for more details.
        \item While we encourage the release of code and data, we understand that this might not be possible, so “No” is an acceptable answer. Papers cannot be rejected simply for not including code, unless this is central to the contribution (e.g., for a new open-source benchmark).
        \item The instructions should contain the exact command and environment needed to run to reproduce the results. See the NeurIPS code and data submission guidelines (\url{https://nips.cc/public/guides/CodeSubmissionPolicy}) for more details.
        \item The authors should provide instructions on data access and preparation, including how to access the raw data, preprocessed data, intermediate data, and generated data, etc.
        \item The authors should provide scripts to reproduce all experimental results for the new proposed method and baselines. If only a subset of experiments are reproducible, they should state which ones are omitted from the script and why.
        \item At submission time, to preserve anonymity, the authors should release anonymized versions (if applicable).
        \item Providing as much information as possible in supplemental material (appended to the paper) is recommended, but including URLs to data and code is permitted.
    \end{itemize}

\item {\bf Experimental Setting/Details}
    \item[] Question: Does the paper specify all the training and test details (e.g., data splits, hyperparameters, how they were chosen, type of optimizer, etc.) necessary to understand the results?
    \item[] Answer: \answerYes{} 
    \item[] Justification: The experimental setting is presented in Section~\ref{sec:experiments} and Appendix~\ref{appendix:evaluation}. 
    \item[] Guidelines:
    \begin{itemize}
        \item The answer NA means that the paper does not include experiments.
        \item The experimental setting should be presented in the core of the paper to a level of detail that is necessary to appreciate the results and make sense of them.
        \item The full details can be provided either with the code, in appendix, or as supplemental material.
    \end{itemize}

\item {\bf Experiment Statistical Significance}
    \item[] Question: Does the paper report error bars suitably and correctly defined or other appropriate information about the statistical significance of the experiments?
    \item[] Answer: \answerNo{} 
    \item[] Justification: Error bars are not reported because it would be too computationally expensive. 
    \item[] Guidelines:
    \begin{itemize}
        \item The answer NA means that the paper does not include experiments.
        \item The authors should answer "Yes" if the results are accompanied by error bars, confidence intervals, or statistical significance tests, at least for the experiments that support the main claims of the paper.
        \item The factors of variability that the error bars are capturing should be clearly stated (for example, train/test split, initialization, random drawing of some parameter, or overall run with given experimental conditions).
        \item The method for calculating the error bars should be explained (closed form formula, call to a library function, bootstrap, etc.)
        \item The assumptions made should be given (e.g., Normally distributed errors).
        \item It should be clear whether the error bar is the standard deviation or the standard error of the mean.
        \item It is OK to report 1-sigma error bars, but one should state it. The authors should preferably report a 2-sigma error bar than state that they have a 96\% CI, if the hypothesis of Normality of errors is not verified.
        \item For asymmetric distributions, the authors should be careful not to show in tables or figures symmetric error bars that would yield results that are out of range (e.g. negative error rates).
        \item If error bars are reported in tables or plots, The authors should explain in the text how they were calculated and reference the corresponding figures or tables in the text.
    \end{itemize}

\item {\bf Experiments Compute Resources}
    \item[] Question: For each experiment, does the paper provide sufficient information on the computer resources (type of compute workers, memory, time of execution) needed to reproduce the experiments?
    \item[] Answer: \answerNo{} 
    \item[] Justification: The experiment did not require too much time. 
    \item[] Guidelines:
    \begin{itemize}
        \item The answer NA means that the paper does not include experiments.
        \item The paper should indicate the type of compute workers CPU or GPU, internal cluster, or cloud provider, including relevant memory and storage.
        \item The paper should provide the amount of compute required for each of the individual experimental runs as well as estimate the total compute. 
        \item The paper should disclose whether the full research project required more compute than the experiments reported in the paper (e.g., preliminary or failed experiments that didn't make it into the paper). 
    \end{itemize}
    
\item {\bf Code Of Ethics}
    \item[] Question: Does the research conducted in the paper conform, in every respect, with the NeurIPS Code of Ethics \url{https://neurips.cc/public/EthicsGuidelines}?
    \item[] Answer: \answerYes{} 
    \item[] Justification: The research adheres to the NeurIPS Code of Ethics. 
    \item[] Guidelines:
    \begin{itemize}
        \item The answer NA means that the authors have not reviewed the NeurIPS Code of Ethics.
        \item If the authors answer No, they should explain the special circumstances that require a deviation from the Code of Ethics.
        \item The authors should make sure to preserve anonymity (e.g., if there is a special consideration due to laws or regulations in their jurisdiction).
    \end{itemize}

\item {\bf Broader Impacts}
    \item[] Question: Does the paper discuss both potential positive societal impacts and negative societal impacts of the work performed?
    \item[] Answer: \answerYes{} 
    \item[] Justification: The potential impacts are discussed in Section~\ref{impacts}. 
    \item[] Guidelines:
    \begin{itemize}
        \item The answer NA means that there is no societal impact of the work performed.
        \item If the authors answer NA or No, they should explain why their work has no societal impact or why the paper does not address societal impact.
        \item Examples of negative societal impacts include potential malicious or unintended uses (e.g., disinformation, generating fake profiles, surveillance), fairness considerations (e.g., deployment of technologies that could make decisions that unfairly impact specific groups), privacy considerations, and security considerations.
        \item The conference expects that many papers will be foundational research and not tied to particular applications, let alone deployments. However, if there is a direct path to any negative applications, the authors should point it out. For example, it is legitimate to point out that an improvement in the quality of generative models could be used to generate deepfakes for disinformation. On the other hand, it is not needed to point out that a generic algorithm for optimizing neural networks could enable people to train models that generate Deepfakes faster.
        \item The authors should consider possible harms that could arise when the technology is being used as intended and functioning correctly, harms that could arise when the technology is being used as intended but gives incorrect results, and harms following from (intentional or unintentional) misuse of the technology.
        \item If there are negative societal impacts, the authors could also discuss possible mitigation strategies (e.g., gated release of models, providing defenses in addition to attacks, mechanisms for monitoring misuse, mechanisms to monitor how a system learns from feedback over time, improving the efficiency and accessibility of ML).
    \end{itemize}
    
\item {\bf Safeguards}
    \item[] Question: Does the paper describe safeguards that have been put in place for responsible release of data or models that have a high risk for misuse (e.g., pretrained language models, image generators, or scraped datasets)?
    \item[] Answer: \answerYes{} 
    \item[] Justification: We will require users to adhere to specific usage guidelines to access the model and datasets, ensuring that they are used responsibly and to mitigate the risk of misuse. 
    \item[] Guidelines:
    \begin{itemize}
        \item The answer NA means that the paper poses no such risks.
        \item Released models that have a high risk for misuse or dual-use should be released with necessary safeguards to allow for controlled use of the model, for example by requiring that users adhere to usage guidelines or restrictions to access the model or implementing safety filters. 
        \item Datasets that have been scraped from the Internet could pose safety risks. The authors should describe how they avoided releasing unsafe images.
        \item We recognize that providing effective safeguards is challenging, and many papers do not require this, but we encourage authors to take this into account and make a best faith effort.
    \end{itemize}

\item {\bf Licenses for existing assets} 
    \item[] Question: Are the creators or original owners of assets (e.g., code, data, models), used in the paper, properly credited and are the license and terms of use explicitly mentioned and properly respected?
    \item[] Answer: \answerYes{} 
    \item[] Justification: We have properly credited the creators or original owners of assets used in the paper and we use the license CC-BY 4.0.
    \item[] Guidelines:
    \begin{itemize}
        \item The answer NA means that the paper does not use existing assets.
        \item The authors should cite the original paper that produced the code package or dataset.
        \item The authors should state which version of the asset is used and, if possible, include a URL.
        \item The name of the license (e.g., CC-BY 4.0) should be included for each asset.
        \item For scraped data from a particular source (e.g., website), the copyright and terms of service of that source should be provided.
        \item If assets are released, the license, copyright information, and terms of use in the package should be provided. For popular datasets, \url{paperswithcode.com/datasets} has curated licenses for some datasets. Their licensing guide can help determine the license of a dataset.
        \item For existing datasets that are re-packaged, both the original license and the license of the derived asset (if it has changed) should be provided.
        \item If this information is not available online, the authors are encouraged to reach out to the asset's creators.
    \end{itemize}

\item {\bf New Assets}
    \item[] Question: Are new assets introduced in the paper well documented and is the documentation provided alongside the assets?
    \item[] Answer: \answerYes{} 
    \item[] Justification: All new assets introduced in this paper will be well documented upon their release. 
    \item[] Guidelines:
    \begin{itemize}
        \item The answer NA means that the paper does not release new assets.
        \item Researchers should communicate the details of the dataset/code/model as part of their submissions via structured templates. This includes details about training, license, limitations, etc. 
        \item The paper should discuss whether and how consent was obtained from people whose asset is used.
        \item At submission time, remember to anonymize your assets (if applicable). You can either create an anonymized URL or include an anonymized zip file.
    \end{itemize}

\item {\bf Crowdsourcing and Research with Human Subjects} 
    \item[] Question: For crowdsourcing experiments and research with human subjects, does the paper include the full text of instructions given to participants and screenshots, if applicable, as well as details about compensation (if any)? 
    \item[] Answer: \answerNA{} 
    \item[] Justification: This paper does not involve crowdsourcing nor research with human subjects.
    \item[] Guidelines:
    \begin{itemize}
        \item The answer NA means that the paper does not involve crowdsourcing nor research with human subjects.
        \item Including this information in the supplemental material is fine, but if the main contribution of the paper involves human subjects, then as much detail as possible should be included in the main paper. 
        \item According to the NeurIPS Code of Ethics, workers involved in data collection, curation, or other labor should be paid at least the minimum wage in the country of the data collector. 
    \end{itemize}

\item {\bf Institutional Review Board (IRB) Approvals or Equivalent for Research with Human Subjects}
    \item[] Question: Does the paper describe potential risks incurred by study participants, whether such risks were disclosed to the subjects, and whether Institutional Review Board (IRB) approvals (or an equivalent approval/review based on the requirements of your country or institution) were obtained?
    \item[] Answer: \answerNA{} 
    \item[] Justification: This paper does not involve crowdsourcing nor research with human subjects.
    \item[] Guidelines:
    \begin{itemize}
        \item The answer NA means that the paper does not involve crowdsourcing nor research with human subjects.
        \item Depending on the country in which research is conducted, IRB approval (or equivalent) may be required for any human subjects research. If you obtained IRB approval, you should clearly state this in the paper. 
        \item We recognize that the procedures for this may vary significantly between institutions and locations, and we expect authors to adhere to the NeurIPS Code of Ethics and the guidelines for their institution. 
        \item For initial submissions, do not include any information that would break anonymity (if applicable), such as the institution conducting the review.
    \end{itemize}

\end{enumerate}


\end{document}